\documentclass[lettersize,journal]{IEEEtran}
\usepackage{amsmath,amsfonts,mathtools,siunitx}
\usepackage{algcompatible}
\usepackage{algorithm}
\usepackage{array}
\usepackage[caption=false,font=normalsize,labelfont=sf,textfont=sf]{subfig}
\usepackage{textcomp}
\usepackage{stfloats}
\usepackage{url}
\usepackage{verbatim}
\usepackage{graphicx}
\usepackage{cite}
\usepackage[nolist]{acronym}
\usepackage{booktabs} 
\newacro{dmp}[DMP]{Dynamic Movement Primitive}
\newacro{seds}[SEDS]{Stable Estimator of Dynamical Systems}
\newacro{gmm}[GMM]{Gaussian Mixture Model}
\newacro{gmr}[GMR]{Gaussian Mixture Regression}
\newacro{gp}[GP]{Gaussian Process}
\newacro{tpgmm}[TP-GMM]{Task-Parameterized \ac{gmm}}
\newacro{lfd}[LfD]{Learning from Demonstration}
\newacro{il}[IL]{Imitation Learning}
\newacro{wls}[WLS]{weighted least-squares}
\newacro{vs}[VS]{Visual Servoing}
\newacro{ds}[DS]{Dynamical System}
\newacro{nn}[NN]{Neural Network}
\newacro{qp}[QP]{Quadratic Problem}
\newacro{ceqln}[CEQLN]{Constrained EQuation Learner Network}
\newacro{eqln}[EQLN]{Equation Learner Network}
\newacro{tpeqln}[TP-EQLN]{Task Parameterized Equation Learner Network}
\newacro{mp}[MP]{Movement Primitive}
\newacro{cnmp}[CNMP]{Conditional Neural Movement Primitive}
\newacro{qop}[QP]{Quadratic optimization Problem}
\newacro{mse}[mse]{Mean of Square Errors}
\newacro{cqr}[CQR]{Constrained Quadratic Regression}

\newacro{bo}[BO]{Bayesian Optimization}
\newacro{bf}[BF]{Basis Functions}
\newacro{sse}[SSE]{Sum-of-Squares Errors}
\newacro{cqp}[CQP]{Constrained Quadratic Problem}
\newacro{sqp}[SQP]{Standard Quadratic Problem}
\newacro{pbd}[PbD]{Programming by Demonstration}
\newacro{crp}[CRP]{Constrained Regression Problem}
\newacro{tpeqln}[TP-EQLN]{Task-Parameterized Equation Learner Networks}
\newacro{cbat}[C-BAT]{Constraint-Based Adapted Trajectories}

\usepackage{todonotes}

\definecolor{darkgreen}{rgb}{0.0,0.49,0.19}

\newcommand{\ecdataset}{{\tilde{\mathcal{D}}}}
\newcommand{\ecy}{{\tilde{\bm{y}}}}
\newcommand{\ect}{{\tilde{\bm{t}}}}
\newcommand{\dmeq}{{\tilde{\bm{\Phi}}}}
\newcommand{\oneseq}{{\tilde{\bm{1}}}}

\newcommand{\yl}{{\underline{\bm{y}}}} 
\newcommand{\yu}{{\bar{\bm{y}}}} 
\newcommand{\tineq}{{\bar{\bm{t}}}} 
\newcommand{\dmineq}{{\bar{\bm{\Phi}}}}
\newcommand{\inecdataset}{{\bar{\mathcal{D}}}}
\newcommand{\onesineq}{{\bar{\bm{1}}}}

\usepackage{authblk}
\usepackage{bm}
\usepackage{multirow}
\usepackage{multicol,lipsum}

\begin{document}

\title{Constrained Equation Learner Networks for Precision-Preserving Extrapolation of Robotic Skills}

\author{Hector Perez-Villeda$^1$, Justus Piater$^{1,2}$, and Matteo Saveriano$^3$
\thanks{$^1$Department of Computer Science, University of Innsbruck, Innsbruck, Austria (E-mail: \texttt{hector.villeda@uibk.ac.at})}
\thanks{$^2$Digital Science Center (DiSC), University of Innsbruck, Innsbruck, Austria (E-mail: \texttt{justus.piater@uibk.ac.at})}
\thanks{$^3$Department of Industrial Engineering, University of Trento, Trento, Italy, (E-mail: \texttt{matteo.saveriano@unitn.it})}
}

\markboth{IEEE Transactions on Robotics,~Vol.~xx, No.~xx, xxx~xxx}{}%


\maketitle

\begin{abstract}
%


In Programming by Demonstration, the robot learns novel skills from human demonstrations of correct executions. After learning, the robot should be able not only to reproduce the skill, but also to generalize it to shifted domains without collecting new training data. Adaptation to similar domains has been investigated in the literature; however, an open problem is how to adapt learned skills to different conditions that are outside of the data distribution, and, more important, how to preserve the precision of the desired adaptations. %
This paper presents a novel supervised learning framework called Constrained Equation Learner Networks that addresses the trajectory adaptation problem in Programming by Demonstrations from a constrained regression perspective.  While conventional approaches for constrained regression use one kind of basis function, e.g., Gaussian, we exploit Equation Learner Networks to learn a set of analytical expressions and use them as basis functions. These basis functions are learned from demonstration with the objective to minimize deviations from the training data while imposing constraints that represent the desired adaptations, like new initial or final points or maintaining the trajectory within given bounds. Our approach addresses three main difficulties in adapting robotic trajectories: 1) minimizing the distortion of the trajectory for new adaptations; 2) preserving the precision of the adaptations; and 3) dealing with the lack of intuition about the structure of basis functions. We validate our approach both in simulation and in real experiments in a set of robotic tasks that require adaptation due to changes in the environment, and we compare obtained results with two existing approaches. Performed experiments show that Constrained Equation Learner Networks outperform state of the art approaches by increasing generalization and adaptability of robotic skills.


\end{abstract}

\begin{IEEEkeywords}
Learning from Demonstration, Learning Basis functions, Constrained Regression,  Trajectory adaptation
\end{IEEEkeywords}
\section{Introduction}\label{sec:intro}

\begin{figure}[t]
    \centering
    \includegraphics[width=\columnwidth]{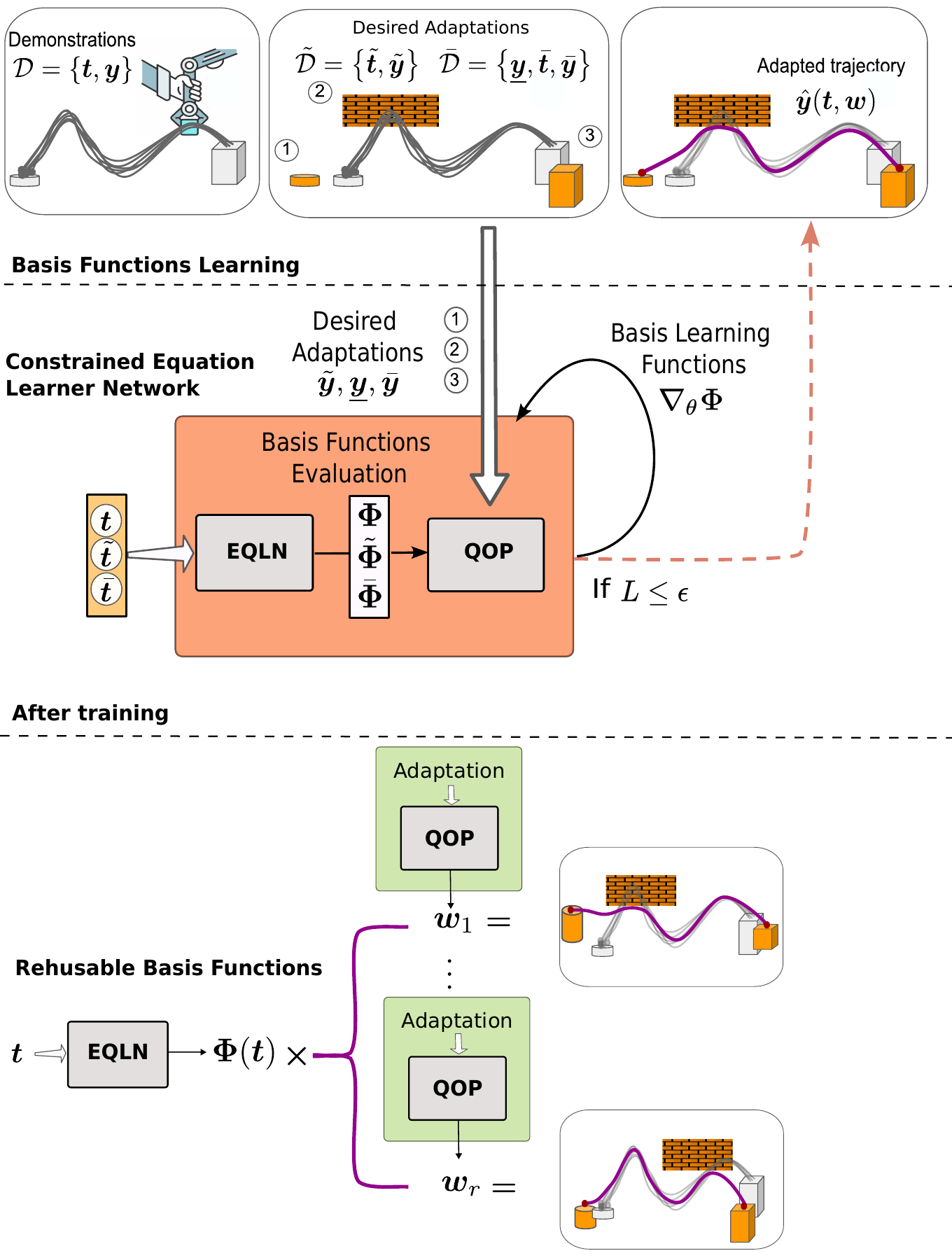}
    \caption{Constrained Equation Learner Networks pipeline. The training data set $\mathcal{D}$ is obtained from a set of demonstrations. The desired adaptation conditions $\bm{y}_A$ and $\bm{y}_l,\bm{y}_u$ are modeled as constraints in a Quadratic Optimization Problem. The Equation Learner Network improves the Basis Functions through backpropagation, aiming to minimize the loss $L$ until it reaches a predefined threshold $\epsilon$. The obtained regressed trajectory $\hat{\bm{y}}$ fulfills the imposed constraints, i.e., the desired adaptation conditions.  Following the training phase, the learned Basis Functions can be reused to recalculate the optimal vector for different new set of adaptations and regress the corresponding trajectory. This flexibility allows for efficient trajectory adaptation without the need for retraining.} 
    \label{fig:CEQLN_apporach}
\end{figure}

\IEEEPARstart{I}{n} the field of machine learning, models are typically trained to perform well on unseen data by generalizing from training samples. However, such models often only work well for interpolation, not extrapolation, meaning that they are ineffective in situations where the data lie outside of the domain they were trained on. Moreover, certain desired properties like precision may be lost when generalizing, especially when extrapolating. These limitations are particularly evident in \ac{pbd} approaches~\cite{billard2016learning,garnelo2018conditional,seker2019conditional,huang2019kernelized, Huang2020Toward,calinon2016tutorial}, which rely on demonstrations and statistical models to encode variations of the important aspects of a task. Such limitations are 1) learned models generalize properly over the training domain, but fail to extrapolate out of this domain; 2) it is necessary to provide demonstrations of the desired adaptation aspect in advance, which can be limiting; and 3) learned models may have some adaptation capabilities, but they lack formal criteria to guarantee that the adaptations are effectively met. This is crucial for tasks that require a guaranteed level of accuracy in adaptation.

To address this problem, we propose to reformulate the trajectory learning as constrained regression, specifically as a constrained \ac{qop}. In this way, we are able to model the desired adaptations as constraints in the regressed space. Imposing the adaptations as constraints gives the possibility to enforce the regressed trajectory to pass by some given points using equality constraints or maintain the values of the trajectory within a given range for a given time slot using inequality constraints. Moreover, we can efficiently regress a novel trajectory solving the \ac{qop}. Overall, by defining the \ac{pbd} problem as Constrained \ac{qop} we can provide close solutions that ensure that the adaptations are fulfilled accurately even for values out-of-the-data distribution while maintaining the spatial relationship of the trajectories~\cite{CQP_Hector}. 

The \ac{qop} is built as a multivariable linear model using \ac{bf}, which allows to get nonlinear models mapping input variables to target variables. 
However, the choice of the appropriate \ac{bf} is a crucial step in the trajectory learning process because it affects the quality of the solution. The use of an inappropriate \ac{bf} can lead to suboptimal solutions, generating significant distortions or even infeasible solutions for the adaptations. This is a significant problem because the \ac{bf} are fixed before the training. A simple solution consists in increasing the number of \acp{bf}. However, the number of \ac{bf} needs to grow exponentially with the dimension of the input, which results in an increased computational cost~\cite{bishop:2006:PRML}. Another possibility is to learn the parameters of the \ac{bf} using stochastic optimization techniques, such as \ac{bo}\cite{BO_2022}. 
However, this method can only handle learning low-dimensional parameters, and it may suffer from convergence problems for high-dimensional parameters. 
We figured out that the problem arises from having a fixed set of \ac{bf}, e.g., Gaussian. Therefore, we exploit the \ac{eqln}~\cite{MartiusL16} to learn the parameters of analytical expressions that are used as basis functions. 
By relying on backpropagation, our approach can overcome the high-dimensional parameter learning problem that arises with stochastic optimization. 

The general pipeline of our approach, called \ac{ceqln}, is shown in Fig.\ref{fig:CEQLN_apporach}. Our method leverages the \ac{eqln} to learn a set of analytical expressions that are used as \ac{bf} for solving a constrained \ac{qop}. In other words, we leave the \ac{eqln} to find the best structure of \ac{bf} by updating the parameters of the activation functions. We use the demonstrations to construct the quadratic function of the \ac{qop}. This ensures that we minimize the distortion of the regressed trajectory and maintain the spatial relationship for trajectories that require adaptations, even for out-of-data distribution domains. On the other hand, the constraints in the \ac{qop} represent the required post-demonstration adaptations of the task due to changes in the environment. The error between the demonstrations and the solution of the constrained \ac{qop} is backpropagated throughout the \ac{eqln} to update its parameters. This approach allows us to efficiently learn the parameters of the basis functions for high-dimensional problems.
%
We extensively validated \ac{ceqln} in simulation and real robot experiments, showing that the proposed approach outperforms the state of the art in out-of-distribution generalization of robotic tasks.


\section{Related Work}\label{sec:related_work}
In the context of \ac{pbd}, a variety of approaches exist that intend to achieve certain level of generalization to different scenarios. We provide a review of existing approaches that specifically focus on task generalization. 

Probabilistic approaches such as Probabilistic \acp{mp}~\cite{paraschos2013probabilistic} provide some generalization by statistical conditioning on the query task parameters, typically a new goal or a set of via-points. Via-point \ac{mp}~\cite{Zhou2019Learning} and Kernelized \ac{mp}~\cite{huang2019kernelized, Huang2020Toward} have better expextrapolation capabilities than Probabilistic \ac{mp}, but also in these approaches task parameters are new goals to reach or a set of via-points to traverse. The \ac{tpgmm}~\cite{calinon2016tutorial} considers as task parameters the homogeneous transformations between arbitrary reference frames. By observing the human demonstrations from each of these frames the robot is able to learn the spatial relationship between start, goal, and via-points in the trajectory. 
Conditional Neural Processes~\cite{garnelo2018conditional} are a class of deep \acp{nn} that combines the function approximation power of \ac{nn} with the data efficiency of Bayesian approaches like \ac{gp}. Inspired by conditional neural processes, Seker et al.~\cite{seker2019conditional} developed an imitation learning framework called \ac{cnmp}. \ac{cnmp} generates motion trajectories by sampling observations from the training data and predicting a conditional distribution over target  points. Training data may include robot position, forces, and any task parameter. All the previously mentioned approaches rely on probabilistic methods to model the data distribution and generate new trajectories by conditioning the learned models. However, these methods have a significant limitation. The new query parameters used to generate new trajectories must fall within the bounds of the data distribution to establish a coherent correspondence between the generated trajectory and the conditioned distribution. This constraint could limit the variety of trajectories that can be generated and may affect the exploration of novel, out-of-distribution trajectories. 

To tackle this problem, a reinforcement learning process was introduced in~\cite{akbulut2020acnmp} to incorporate out-of-distribution information. By doing so, it becomes feasible to extend the data distribution boundaries and produce trajectories for query parameter values that lie beyond the initial data distribution.
On the other hand, in our previous work \cite{Villeda-2022-RAS}, we developed a new method called \ac{tpeqln}. \ac{tpeqln} extend the \ac{eqln}~\cite{MartiusL16} with task parameters for learning and generalizing robot skills. \ac{tpeqln} was compared againts \ac{tpgmm} and \ac{cnmp}, showing superior performance especially when generalizing beyond the domain of the training data. 
However, \ac{tpeqln} has some limitations: 1) noisy demonstrations or not well structured can affect the performance of the method in extrapolation; 2) the further the query values in extrapolation domain, the lower is the performance; 3) there are not close solutions that guaranties to fulfill the required adaptations of the trajectory related to the feature parameter. As an attempt to mitigate these problems we developed a second approach \cite{CQP_Hector}. Here, we address the problem from the  constrained regression perspective, where the set of demonstrations are used as target values to build a \ac{qop}, and the desired adaptations of the trajectory are represented as either  equality or inequality constraints. The approach guarantee of fulfilling the required adaptations in a close-form, even when they extend beyond the boundaries of the demonstration data distribution. However, one of the limitations is that the approach critically depends on the selection of appropriate \acp{bf}. If the \acp{bf} are not well-suited, the method may introduce distortions in the regressed trajectory or lead to not feasible solutions of the \ac{qop}.

The limitation of the presented approaches, specially from the \ac{tpeqln} and \ac{cbat} are effectively overcome by the proposed \ac{ceqln}.  


The selection of \ac{eqln} as a method to fit the \acp{bf} is based on its demonstrated superior generalization performance, even in extrapolation scenarios. The introduction of a set of diverse elemental functions as a part of a regression problem is not new and it has been already explored in varios context such as identification of dynamical systems \cite{Brunton_2016, BRUNTON2016710, Lejarza_2022}.

Recent research has combined trajectory optimization with learning from demonstrations to enhance robot performance. Shyam et al.~\cite{shyam2019improving} propose to generate initial solutions for local optimization by estimating a trajectory distribution from expert demonstrations. Huang~\cite{huang2021ekmp} extended kernelized movement primitives (EKMP) that learn probabilistic features, adapt skills towards desired points, avoid obstacles, and satisfy constraints.  Rana et al. \cite{rana17towards} presented Combined Learning from Demonstration And Motion Planning, which unifies learning from demonstration and motion planning using probabilistic inference to find optimal and feasible trajectories for different scenarios.  However, these approaches may struggle to handle accurate adaptations for the demonstrated task since the methods are based on probabilistic and trajectory distribution. In \ac{ceqln}, we tackle this problem by modeling the desired adaptations with equality constraints, and providing close solutions for the adaptation, enforcing the trajectory to converge to the desired points.

Saveriano et al.~ \cite{saveriano2019learning}, developed an approach to plan robotic motions in real-time using constraints learned from human demonstrations and generating motion trajectories that stay within the constrained workspace. In \cite{ames2019control}, the authors use control barrier functions  to verify and enforce safety properties in the context of optimization based safety-critical controllers. Although these approaches offers a stable dynamic system that adapt the trajectory to novel situations, the adaptations might also suffer of lack of criteria to  maintain the shape of the demonstrated trajectory for the adaptations. In \ac{ceqln}, we tackle this problem by using the demonstrations to build the quadratic functions used in the \ac{qop} and in this way minimize the distortion of the regressed trajectory.
 
Osa et al. \cite{osa2017guiding} tackle the distortion of the adapted trajectory by proposing a motion planning framework that learns a distribution of demonstrated trajectories to guide trajectory optimization while adapting the trajectory to avoid obstacles. In \ac{ceqln}, the obstacle avoidance is modeled as inequality constraint and included in the \ac{qop}, this allows to manage multiple optimization criteria at the same time, for example, avoid the obstacle, while adapting the initial and final point of the trajectory and minimizing the distortion of the trajectory at the same time.  Alternative methods, proposed in \cite{vochten2018invariant, VOCHTEN2019103291}, focus on preserving the shape of the adapted trajectory by using invariant-based techniques. Perico et al. \cite{perico2019combining} developed a method that combines imitation learning with model-based and constraint-based task specification and control. They used a statistical uni-modal model to describe demonstrations and combined it with model-based descriptions of the task.  These approaches present some similarities with \ac{ceqln} regarding  the 
representation of the demonstrations in terms of a number of weighted basis functions. However, as we demonstrate along this paper, the use of fixed basis functions has some limitations, for example, they might not be the right ones for the given regression problem, requiring a manual tuning process of their parameters to improve the results.  In \ac{ceqln}, we overcome this problem by introducing a supervised learning process where the best parameters of the basis functions are learned. 

Other ways to address the challenge of generalizing demonstrated trajectories have been explored under the concept of null-space projection. Lin et al. \cite{lin2017learning} developed a method for learning self-imposed constraints in movement observations using the operational space control framework. The approach estimates the constraint matrix and its null space projection to decompose the task space and any redundant degree of freedom. The method requires no prior knowledge of the dimensionality of the constraints or the control policies. In the same direction, Manavalan et al. \cite{manavalan2020library} introduced an open source software library called Constraint Consistent Learning (CCL) that implements a family of data-driven methods. The library can learn state-independent and -dependent constraints, decompose the behavior of redundant systems into task- and null-space parts, and uncover the underlying null space control policy. While these approaches use fixed \acp{bf} and demonstrate promising results, the potential for enhanced generalization remains open. Our method introduces a supervised learning process to refine the set of \acp{bf}, leading to improved generalization, a distinction from these prior methods. A comparable trajectory generalization idea is presented by Nordmann et al. \cite{nordman2012teaching}, where authors leverage null-space constraints learned through a reservoir neural network for human-robot interaction. The integration of null-space constraints is a direction we consider for future exploration with \ac{ceqln}. This exploration could be to learn appropriate \acs{bf} to learn null-space projections matrix  while fulfilling imposed task-space constraints.

Frank et al. \cite{frank2021constrained} created a probabilistic framework for adapting probabilistic movement primitives (ProMPs) with constrained optimization. Additionally, Jankowski et al. \cite{Jankowski22RAL} investigate the synergy between learning adaptive movement primitives and key position demonstrations. They exploit a linear optimal control formulation to recover the timing information of the skill missing from key position demonstrations and to infer low-effort movements on-the-fly.
Vochten et al. \cite{Vochten2021} present an  approach for adapting end-effector trajectories in robot manipulators while preserving their original motion characteristics. The method employs a coordinate-invariant shape description and integrates with a reactive control framework. Vergara et al.\cite{inproceedings2020Vergara_Perico} developed a framework that combines learning from demonstration and constraint-based task control to enhance robotic automation. It tackles high variability and uncertainty by adapting approach motions based on learned information and sensor updates. Vochten et al.\cite{VochtenIROS2018} provides a new approach for enhancing the accuracy of invariant representations of demonstrated motion trajectories. They introduces an optimization-based method that minimizes the error between the reconstructed trajectory from the invariant representation and the measured trajectory. Manavalan et al. \cite{Burlizzi9982222} combine trajectory-parameterized Probabilistic Principal Component Analysis (traPPCA) with the invariants method to enhance generalization in Learning from Demonstration using virtual demonstrations. In contrast to our approach, they utilize Model Predictive Control for new virtual demonstrations. Our method enables direct adaptation from generalized demonstration space using learned basis functions (\acp{bf}). While their method suggests potential for adaptations to initial and final trajectory points, as well as inclusion of inequality constraints in Model Predictive Control, these aspects are not reported in any case studies. Although these approaches tackle the adaptation from the \ac{pbd} and constraint perspective, they lacks consideration for inequality constraints, which are very useful to achieve more complex behaviour in the task, such as obstacle avoidance or maintain the adapted trajectory in a specified range. Additionally, the trajectory generated using the previous approaches highly depends on the correct choose of the parameters and weights in the optimization process. This raises a pertinent question of how those parameters should be selected. 

On the other hand, \ac{ceqln} is a novel method that take into considerations constraints into the optimization process to adapt and generalize trajectories in robotics context. Some efforts to integrate constraints into the learning process of Neural Networks have been applied already.
In the context of physics informed deep learning, ~\cite{daw2021physicsguided, jia2019physics, raissi2018hidden} introduce a loss-constrained \ac{nn}, where the introduction of soft constraints are implemented through a "penalty factor" is incorporated into loss function. The penalty factor quantifies the deviation of the output from each of the  constraints, and this penalty is then included in the cost function that the network aims to minimize. However, this way does not provide close solutions to the constraint conditions.

Beucler et al. \cite{beucler2021enforcing} incorporate constraints into a \ac{nn} in a different way. They enforce $n$ constraints by introducing an equal number of "residual" layers within the \ac{nn} to guarantee that model meet multiple physical laws. A key difference from \ac{ceqln} lies in the choice of the quadratic function for minimization. While \cite{beucler2021enforcing} uses the MSE as the quadratic function to minimize, \ac{ceqln} employs the quadratic standard form evaluation of the \acp{bf}. This choice comes with distinct advantages.  The main one is that, once the parameters of the \acp{bf} are fitted, they can be consistently applied to regress trajectories under new sets of constraints. This feature significantly enhances the generalization capabilities of our method.


All previously mentioned methodologies share some common ground with our approach, particularly in the aspects concerning parameter and basis function selection. These overlaps opens the door for these methodologies to leverage our approach's advantages. For instance, they could incorporate our method to compute \acp{bf}, benefiting from the added capability of constraints inclusion. This presents potential research avenues to explore and opportunities for improving in this field.


\section{Problem Definition}\label{sec:prob_definition}  
Using a \ac{qop} formulation to impose hard constrains on the regressed trajectory can potentially cause large deviations from the desired path. In some cases, the nature of the constraints itself imposes these deviations to correctly execute the task. However, in many other cases, we found out that the accuracy of the \ac{qop} is strongly affected by the choice of the \ac{bf}. 



\begin{figure}[t]
    \centering
    \includegraphics[width=\columnwidth]{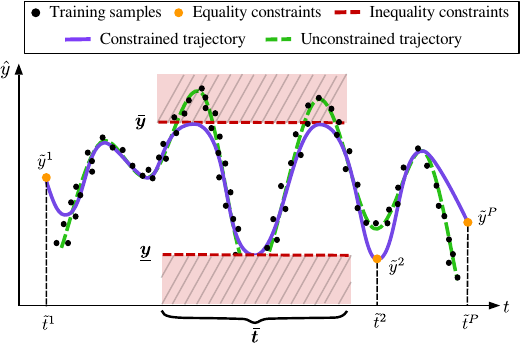}
    \caption{Constrained regression problem. The training data set $\mathcal{D}$ is used to create the regression model. The equality constraint data set $\tilde{\mathcal{D}}$ is used to enforce the regressed trajectory $\hat{\bm{y}}$ to pass by the points $\ecy$ at $\ect$, whereas the inequality constraints data set $\inecdataset$ is used to maintain the values of $\hat{\bm{y}}$ within the boundaries $\yl$ and $\yu$ for $\tineq$ steps.} 
    \label{fig:constrained_reg}
\end{figure}

\begin{figure}[t]
    \centering
    \includegraphics[width=1\columnwidth]{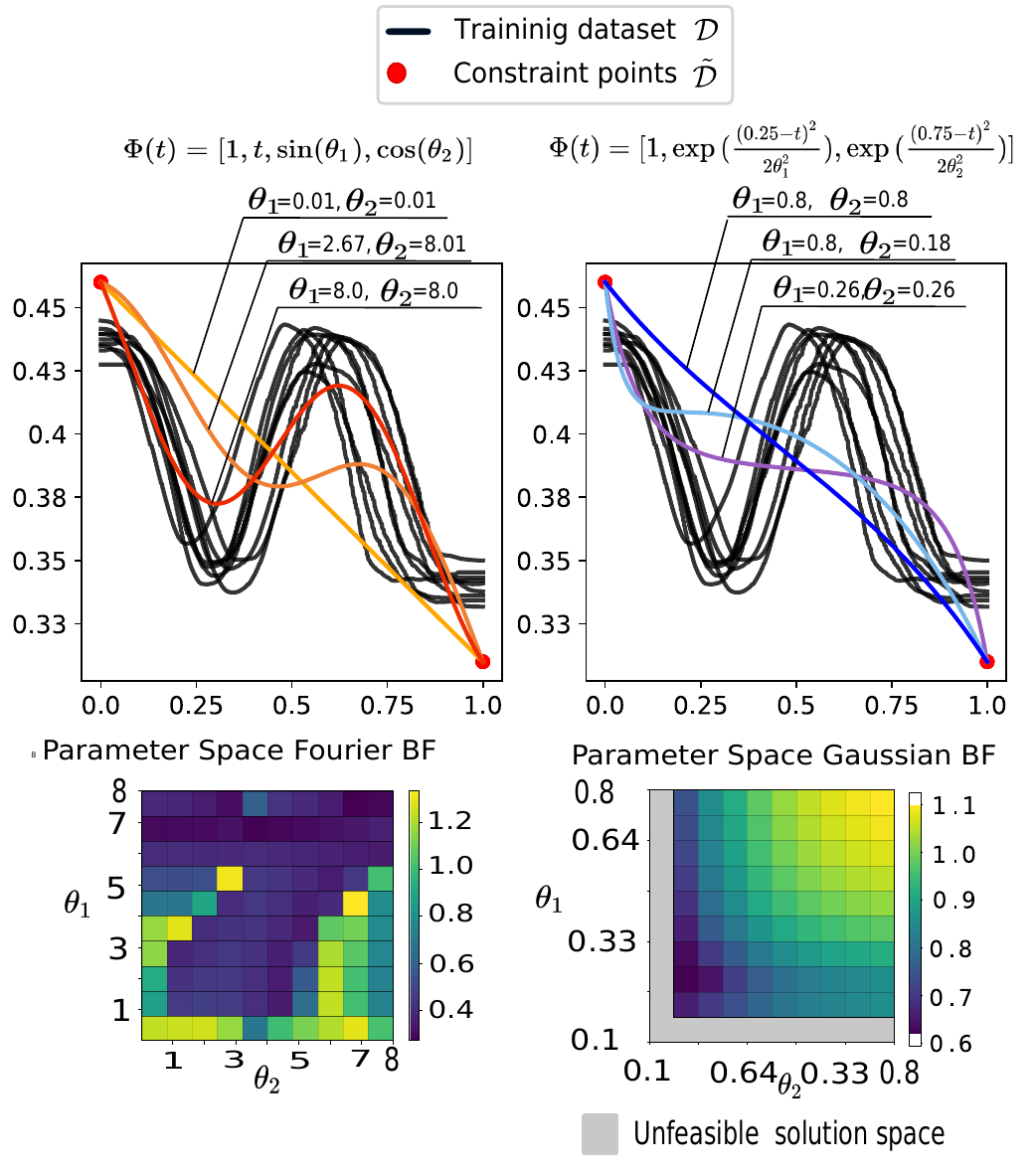}
    \caption{Toy example for constrained regression in $1$D. (Top) Three trajectories are depicted to show the impact of the model parameters of $\theta_1$ and $\theta_2$. The training data set $\mathcal{D}$ is used to construct the quadratic function of the \ac{qop} and the equality constraints data set $\ecdataset$ is used to adapt the regressed trajectory to the desired way points.  (Bottom) \ac{sse} obtained for each value of $\theta_1$ and $\theta_2$. The plot shows the impact of the model parameters in the constrained regression problem.} 
    \label{fig:1D_toy_example}
\end{figure}

To give a better intuition about the problem we are tackling, we present a $1$D toy example to show the importance of the parameter selection of $\bm{\Phi}$ and its limitations in a constrained regression problem. We consider a data set $\mathcal{D}=\left \{ \bm{t},\bm{y} \right \}$ consisting of $3500$ data points where the $\bm{t}$ vector
is equally distributed within $0\leq \bm{t}\leq 1$. We also define a set of desired adaptations $\ecdataset=\left \{\ect,\ecy \right \}$ with $\ecy=[0.46,0.31]^\top$ for the initial and final point, located at $\ect=[0,1]^\top$ respectively. The desired adaptations are modeled as 
equality constraints in~\eqref{eq:cqp}. 
The training and constraint data sets are depicted in Fig.~\ref{fig:1D_toy_example} (top). We tested two set of  \acp{bf} used in the literature, i.e., Fourier and Gaussian \acp{bf}. 


As Fourier \acp{bf} we use
\begin{equation}\label{eq:Fourier_bf_toy_example}
\Phi(t)=[1,t,\sin(\theta_1 t),\cos(\theta_2 t)]^{\top},
\end{equation}
where $\left \{\theta_1, \theta_2\right \}$ are the parameters to be chosen.
As Gaussian \acp{bf} we use
\begin{equation}\label{eq:Gaussianbf_toy_example}
\Phi(t)=\left[1,\exp{\left(\frac{0.25-t^2}{2\theta_1} \right)},\exp{\left(\frac{0.75-t^2}{2\theta_2}\right)}\right]^{\top},
\end{equation}
where $\left \{\theta_1, \theta_2\right \}$ are the parameters to be chosen. In this case, we have fixed the Gaussians at $0.25$ and $0.75$, and varied the width of each Gaussian by using the parameters $\theta_1$ and $\theta_2$, respectively. It should be noted that the number of \acp{bf} used is limited, so the resulting regressed trajectory may have a relatively large \ac{mse} (the larger the \ac{mse}, the greater the distortion of the regressed trajectory). However, the main goal is to demonstrate the performance variations with respect to the parameter values. We have chosen to tune only two parameters, $\theta_1$ and $\theta_2$, so that the \ac{mse} can be visualized in a two-dimensional parameter space. 

To show the impact of parameter selection on the performance of the proposed method, we vary the values of $\theta_1$ and $\theta_2$ independently and linearly. Specifically, for the first case (Fig.~\ref{fig:1D_toy_example} (top-left)), we vary the parameters in the range $\theta_1,\theta_2=0.01+0.799k$, with $k=0,\ldots,9$. For each value of $k$, we solve the \ac{qop} in~\ref{eq:cqp} using the \acp{bf} in~\eqref{eq:Fourier_bf_toy_example}, obtaining $100$ solutions. 
%
In Fig.\ref{fig:1D_toy_example} (top-left), we show the trajectories regressed using $\theta_1$ and $\theta_2$ that lead to the best, intermediate, and worst \ac{mse}. The MSE obtained from the $100$ parameter variations is shown in Fig.~\ref{fig:1D_toy_example} (bottom-right). It can be seen that all the trajectories satisfy the equality constraints. However, the choice of $\theta_1$ and $\theta_2$ has a significant impact on the reproduction accuracy.  

In the second case, we also vary the parameters as $\theta_1,\theta_2=0.1+0.07k$ 
with $k=0,\ldots,9$. We solve the \ac{qop} for each parameter set with the Gaussian \acp{bf} in~\eqref{eq:Gaussianbf_toy_example}. In Fig.\ref{fig:1D_toy_example} (top-right), we show the trajectories regressed using $\theta_1$ and $\theta_2$ that lead to the best, intermediate, and worst \ac{mse}. The MSE obtained from the $100$ parameter variations is shown in Fig.~\ref{fig:1D_toy_example} (bottom-right). It can be seen that all the trajectories satisfy the equality constraints. Also in this case, the choice of $\theta_1$ and $\theta_2$ has a significant impact on the reproduction accuracy. Note that the gray area in the plot represents unfeasible solutions.


In conclusion, while a \ac{qop} formulation is effective in fulfilling the constraints, we observed that the value of the parameters $\theta_1$ and $\theta_2$ directly influences the accuracy of the regressed trajectory. 
In robot learning, it is not only important to fulfill the adaptation constraints, but also to preserve the shape of the trajectory which is a relevant aspect to ensure the successful completion of the task. 
Additionally, in the second case, we also observe that for some parameter values there could be unfeasible solutions for the \ac{qop}. This highly depends on the structure of $\bf{\Phi}$ and it highlights the importance of considering the structure of the \acp{bf} for $\bf{\Phi}$.
Our solution to these limitations, namely the \ac{ceqln}, is described in the following section.

\section{Method}\label{sec:method}
In this section, we provide a detailed explanation of the the proposed \acf{ceqln}. We first describe the network architecture, then the strategy used to train it. 


\subsection{The \ac{ceqln} architecture}
As shown in Fig.~\ref{fig:CEQLN_architecture}, the architecture of the \ac{ceqln} is composed of four main blocks. 
\subsubsection*{\textbf{Block 1}} is an \ac{eqln}~\cite{MartiusL16}, which is a multi-layered feed-forward network with $L$ layers: $l=\left \{1, \ldots, L-1\right \}$  are hidden layers and the last layer $L$ is an output layer. The input  $\bm{z}^{l}$ to the $l$-th layer, which is the output of the $l-1$-th layer, is defined as
\begin{equation}\label{eq:z}
    \bm{z}^{(l)} = \bm{W}^{(l)}\bm{h}^{(l-1)}+\bm{b}^{(l)} = \bm{W}^{(l)}\bm{f}\left({\bm{z}^{(l-1)}}\right)+\bm{b}^{(l)},
\end{equation}
where $\bm{h}^{(l-1)}$ is the output of the preceding layer $l-1$, $\bm{W}^{(l)}$ is a matrix of weights, and $\bm{b}^{(l)}$ is a bias vector. $\bm{h}^{(0)}$ and $\bm{h}^{(L)}$ are defined as the network input and output respectively. $\bm{W}$ and $\bm{b}$ are the open parameters to learn that we will define as $\bm{\theta}=\left \{\bm{W},\bm{b}\right \}$. While a traditional \ac{nn} uses one kind of activation function $\bm{f}$, the \ac{eqln} uses the following activation functions: 
\begin{equation}
\begin{array}{rclcl}
f_0(z)&=& I(z) &=& z,\\ 
f_1(z)&=&\sin(z),\\
f_2(z)&=&\cos(z),\\
f_3(z)&=&\sigma(z) &=& \dfrac{1}{1+\text{e}^{-z}},\\
f_4(z_{0},z_{1})&=&z_{0} z_{1},\\ 
f_5(z)&=& \text{sech}(z) &=& \dfrac{2}{\text{e}^{z}+\text{e}^{-z}}.\\
\end{array}
\label{eq:activation_functions}
\end{equation}
For a multidimensional input the above activation functions are applied element-wise.
The output $\bm{h}^{(L)}$ is a fully-connected layer with linear activation functions that, in our approach, represent the set of \acp{bf} $\{\phi_1,\ldots,\phi_M\}$ estimated at each time step $t$. Each $\phi_m$, $m=1,\ldots,M$, depends on the parameters of the \ac{eqln} $\bm{\theta}$ which are updated using backpropagation. 
%
%
%
\begin{figure*}[h]
    \centering
    \includegraphics[width=1.0\textwidth]{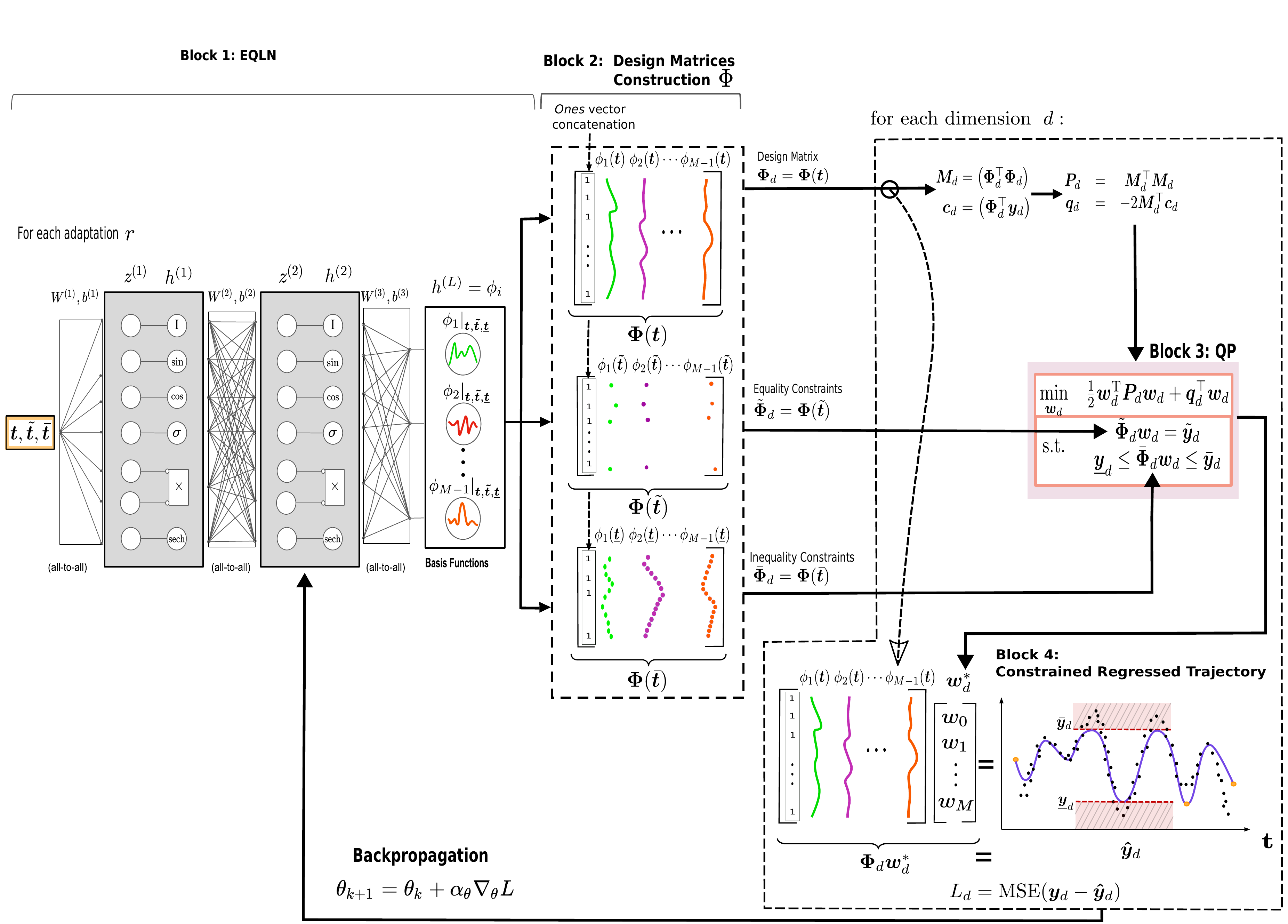}
    \caption{The \ac{ceqln} \textbf{pipeline}. The architecture consists of four main blocks. \textbf{Block 1} evaluates the \ac{eqln} with each time step vector of the data set $\mathcal{D}$, $\ecdataset$, and $\inecdataset$. \textbf{Block 2} constructs the design matrices for each evaluation. \textbf{Block 3} constructs and solves the \ac{qop}, and \textbf{Block 4} calculates the trajectory and its cost function. This process is repeated for the entire set of constraints $r$. The total loss function is then used to calculate the gradient of the parameters $\theta$ of the \ac{eqln} and update them.}
    \label{fig:CEQLN_architecture}
\end{figure*}
The \ac{eqln} architecture used in this work is depicted in Fig.~\ref{fig:CEQLN_architecture}~(\textbf{Block 1}). The \ac{eqln} learns in a supervised way a proper set of \ac{bf}. Indeed, the last layer of the \ac{eqln} has $M-1$ outputs, representing the evaluations of the \acp{bf}. Here, the time vectors of each data set $\bm{t}$, $\ect$, $\tineq$ are evaluated in the \ac{eqln}, to obtain  $\phi_i(\bm{t})$, $\tilde{\phi}_i = \phi_i(\ect)$, and $\bar{\phi}_i = \phi_i(\tineq)$ respectively.

 

\subsubsection*{\textbf{Block 2}} rearranges the evaluations to compute the design matrices used in the \ac{qop} defined in~\eqref{eq:cqp}. In particular, $\phi_i$ is concatenated element-wise with a vector of ones $\bm{1}\in\mathbb{R}^{\mathrm{N}}$ to construct the design matrix $\bm{\Phi}$ defined in~\eqref{eq:Design_Matrix}. Similarly, $\tilde{\phi}_i$ is concatenated element-wise with the vector $\oneseq\in\mathbb{R}^{\mathrm{P}}$ and $\bar{\phi}_i$ with $\onesineq\in\mathbb{R}^{\mathrm{Q}}$ to construct the design matrices $\tilde{\bm{\Phi}}$ and $\bar{\bm{\Phi}}$ defined in~\eqref{eq:cqp}.
 
\subsubsection*{\textbf{Block 3}} defines and solves a \ac{qop} used to formulate a constrained regression problem~\cite{CQP_Hector}.
In our setting, we assume that a set of $N$ observations $\mathcal{D}=\left\{t^n,\bm{y}^n\right \}_{n=1}^N$ is given, where $t^n\in\mathbb{R}$ defines the independent variable (e.g., the time) and $\bm{y}^n\in\mathbb{R}^{D}$ the target values of dimension $D$ (e.g., the robot pose). 
Let's first consider the case where the dependent variable is a scalar, i.e., $\mathcal{D}=\left\{t^n,y^n\right \}_{n=1}^N$, the general case will be described later in this section. 

Given $\mathcal{D}$, we can fit a multivariate linear model in the form
\begin{equation}\label{eq:Estimated_model}
\hat{\bm{y}}(t)= w_0 + \sum_{m=1}^{M-1} \phi_m(t) w_m = \bm{w}^{\top}\bm{\phi}(t),
\end{equation} 
where $\bm{\phi} = [1, \phi_1, \ldots, \phi_{M-1}]^{\top}$ is a vector of nonlinear \ac{bf} (e.g., Gaussians) depending on $t$, and $\bm{w} = [w_0,\ldots,w_{M-1}]^{\top}$ is a vector of weights computed by minimizing the regularized \ac{sse}
\begin{equation}\label{eq:sum-of-squares}
L=\frac{1}{2}\sum_{n=1}^{N} \left(\bm{w}^{\top}\bm{\phi}(t^n) - y^n \right)^{2}+\frac{\lambda_w}{2}\Vert \bm{w} \Vert^2 .
\end{equation}
The regression problem~\eqref{eq:sum-of-squares} can be rewritten as a \ac{qp} in the form
\begin{equation}\label{eq:qp}
\underset{\bm{w}}{\text{minimize}} \quad \frac{1}{2}\bm{w}^{\top}\bm{P}\bm{w} + \bm{q}^{\top}\bm{w},
\end{equation}
where $\bm{P} =2\bm{M}^{\top}\bm{M}+\lambda_w\bm{I}$, $\bm{q} = -2\bm{M}^{\top}\bm{c}$, $\bm{M} = (\bm{\Phi}^\top\bm{\Phi})$ and $\bm{c} = \bm{\Phi}^\top\bm{y}$. The  \textit{design matrix} $\bm{\Phi}\in\mathbb{R}^{N\times M }$ is defined as
\begin{equation}\label{eq:Design_Matrix}
\bm{\Phi}(\bm{t})=\begin{bmatrix}
1 &  \phi_1(t^1)  & ... & \phi_{M-1}(t^1)   \\ 
1 &  \phi_{1}(t^2)   & ... &  \phi_{M-1}(t^2) \\ 
\vdots &  \vdots  & \vdots  & \vdots    \\ 
1 &   \phi_{1}(t^N)  & ... &  \phi_{M-1}(t^N) \\ 
\end{bmatrix},
\end{equation}
where each row represents the evaluation of the set of \ac{bf} for a given time step $t^n$.

By defining the regression problem as \ac{qop}, it becomes possible to introduce both equality and inequality constraints in the regression space. 
This allows for precise control over the shape of the regressed trajectory, such as ensuring that it passes through specific waypoints or maintains certain values within a given range for a specified duration. This flexibility is particularly valuable in a wide range of applications. The constrained \ac{qp} is formulated as
\begin{equation}
\begin{aligned}
& \underset{\bm{w}}{\text{minimize}}
& & \frac{1}{2}\bm{w}^{\top}\bm{P}\bm{w} + \bm{q}^{\top}\bm{w} \\
& \text{subject to}
& &\dmeq\bm{w}=\ecy, \\
&&& \yl\leq \dmineq\bm{w}\leq\yu.
\end{aligned}
\label{eq:cqp}
\end{equation}
The formulation of the constraints in equation \eqref{eq:cqp} enables us to impose desired adaptations on the regressed trajectory $\hat{\bm{y}}$ while optimizing the cost function. This can be visualized in Fig.~\ref{fig:constrained_reg}, which illustrates a simple constrained regression problem. On one hand, we have equality constraints defined in a data set $\tilde{\mathcal{D}}=\left \{\tilde{t}^p,\tilde{y}^p \right \}_{p=1}^P$ with $P \leq N$ steps, that ensure that the regressed trajectory $\hat{\bm{y}} = [\hat{y}(t^1),\ldots,\hat{y}(t^N)]^{\top}$ passes through the orange points $\ecy = [\tilde{y}(t^1),\ldots,\tilde{y}(t^P)]^{\top}$ at the corresponding steps $\ect = [\tilde{t}^1,\ldots,\tilde{t}^P]^{\top}$.  The equality constraints are formulated as $\dmeq\bm{w}=\ecy$ where  $\dmeq = \bm{\Phi}(\ect)\in\mathbb{R}^{P\times M }$.

Similarly, the inequality constraints defined in a data set  $\bar{\mathcal{D}}=\left \{\underline{y}^q,\bar{y}^q,\bar{t}^q \right \}_{q=1}^Q$ with $Q \leq N$ steps allows to maintain the values of the regressed trajectory within the range $\yl\leq \hat{\bm{y}}\leq \yu$ at the corresponding steps $ \tineq = [\bar{t}^1,\ldots,\bar{t}^Q]^{\top}$ (dashed red line in Fig.~\ref{fig:constrained_reg}). The inequality constraints are formulated as $\yl\leq \dmineq\bm{w}\leq\bar{ \bm{y}}$, where $\yl = [\underline{y}^1,\ldots,\underline{y}^Q]^{\top}$, $\bar{ \bm{y}} = [\bar{y}^1,\ldots,\bar{y}^Q]^{\top}$, and $ \dmineq=\bm{\Phi}( \tineq)\in\mathbb{R}^{Q\times M }$.

The \ac{qop} formulation in~\eqref{eq:cqp} can be extended to multi-dimensional data sets following the same intuition. Let us consider the set of $N$ observations $\mathcal{D}=\left\{t_n,\bm{y}_n\right \}_{n=1}^N$, where each observation is a vector of $D$ elements, i.e., $\bm{y}_n\in\mathbb{R}^{D}$.
The quadratic cost in~\eqref{eq:cqp} remains formally the same in this case by redefining  $\bm{M}$ as a block diagonal matrix in the form $\bm{M}=\mathrm{blkdiag}(\bm{M}_1,\ldots,\bm{M}_D)\in\mathbb{R}^{DM\times DM}$, where each $\bm{M}_d = \bm{\Phi}^{\top}\bm{\Phi}$. The weights are stacked in the matrix $\bm{w}=[{\bm{w}_1^\top}, \cdots, {\bm{w}_D^\top}]^\top\in\mathbb{R}^{DM}$ and, similarly, $\bm{c}=[{\bm{c}_1^\top}, \cdots, {\bm{c}_D^\top}]^\top\in\mathbb{R}^{DM}$.
In the same way, the equality constraints in~\eqref{eq:cqp}  are redefined as $\dmeq=\mathrm{blkdiag}(\dmeq_1,\ldots,\dmeq_D)\in\mathbb{R}^{DP\times DM}$ and $\ecy=[{\ecy_{1}^{\top}}, \cdots, {\ecy_{D}^{\top}}]^{\top}\in\mathbb{R}^{DP}$, while inequality constraints become $\dmineq=\mathrm{blkdiag}(\dmineq_1,\ldots,\dmineq_D)\in\mathbb{R}^{DQ\times DM }$, $\yl=[\yl_{1}^\top, \cdots, {\yl_{D}^\top}]^\top\in\mathbb{R}^{DQ}$, and  $\yu=[\yu_1^\top, \cdots, \yu_D^\top]^\top\in\mathbb{R}^{DQ}$. 

It is also possible to extend the definition of equality and inequality constraints to multiple sets of constraints $r\in\left \{1, \ldots,R  \right \}$, generating  different equality and inequality constraints data sets $\ecdataset^r$ and $\inecdataset^r$ respectively. For each set of constraints $r$, we can construct a different \ac{qop}, resulting in a different set of optimal variables $\bm{w}$ that regress a different trajectory $\hat{\bm{y}}$ that  fulfill specifically the constraint $r$ but using the same \acp{bf}. In the context of \ac{pbd} and trajectory regression, each of this set of constraints represent different adaptations of the task.


To summarize, in a multi-dimensional case, the construction of the \ac{qop} involves calculating the matrices $\bm{M}_d$ and the vector $\bm{c}_d$ for the cost function, as well as $\dmeq_d$ and $\dmineq_d$ for the constraint matrices, for each dimension $d$. Once this step is completed, the final matrices $\bm{M}$, $\dmeq$, and $\dmineq$, along with the final vectors $\bm{c}$, $\ecy$, $\yl$, and $\yu$, can be constructed for the multidimensional case.
 Finally, the matrix $\bm{P}$ and the vector $\bm{c}$ are computed  to construct the quadratic function as well as the equality and inequality constraints $\dmeq\bm{w}=\ecy$,  $\yl\leq\dmineq\bm{w}\leq\yu$ respectively. At this point, the \ac{qop}  defined in~\eqref{eq:cqp} is solved to obtain the optimal vector $\bm{w^{*}}\in\mathbb{R}^{DM}$. 
 
\subsubsection*{\textbf{Block 4}} uses the optimal weights $\bm{w^{*}}$ and the design matrix  $\bm{\Phi}$ to compute the regressed trajectory $\hat{\bm{y}} \in \mathbb{R}^{D \times N}$ is computed as $\hat{\bm{y}} =[\bm{\Phi}\bm{w}^{*}_1,\ldots,\bm{\Phi}\bm{w}^{*}_D]^{\top}$. The regressed trajectory is used to compute the \ac{sse} that is then backpropagated to update the parameters $\bm{\theta}$ of the \ac{eqln} in \textbf{Block 1}. With the new set of parameters,  the \ac{eqln} generates a new set of \acp{bf} resulting in a new regressed trajectory. The objective of the training is to accurately reproduce the demonstrations while fulfilling the constraints. The training procedure for the \ac{ceqln} is detailed as follows.   

\subsection{Training the \ac{ceqln}}
\begin{algorithm}[t]
\caption{\ac{ceqln} Training}\label{alg:CEQLN_training}
\begin{algorithmic}[1]
\REQUIRE $\mathcal{D}=\left \{ \bm{t},\bm{y} \right \}$, $\ecdataset^r=\left \{ \ect^r,\ecy^r \right \}$, $\inecdataset^r=\left \{ \tineq^r,\yl^r,\yu^r \right \}$, $[\theta_a,\theta_b]$, $\varepsilon$, $\beta$
\STATEx \textbf{EQLN Initialization}
\FOR {$k \gets 1$ to $\beta$} 
\STATE $\bm{\Phi}_\theta\gets \theta \sim U(\theta_a,\,\theta_b)$
\FOR {$r \gets 1$ to $R$}
\STATE $\mathcal{D}\gets\left \{ \bm{t},\bm{y} \right \}$,  $\ecdataset\gets\left \{\ect,\ecy \right \}$, $\inecdataset=\left \{ \tineq,\yl,\yu \right \}$
\STATE $\hat{\bm{y}}\gets$\texttt{\ac{ceqln}\_evaluation}$(\mathcal{D},\ecdataset,\inecdataset)$ 
\STATE $L=L+$\texttt{\ac{sse}}$(\hat{\bm{y}_d},\bm{y}_d)$
\ENDFOR
\STATE $\mathcal{G}\gets\left \{\theta,L\right \}$
\ENDFOR 
\STATEx \textbf{Training Process}
\STATE $\bm{\Phi}_\theta\gets\underset{\mathbf{\theta}}{\text{argmin}}\hspace{0.2cm} \mathcal{G}$
\STATE\textbf{for  }$k \gets 1$ to $\epsilon$ \textbf{do}
\STATE\hspace{0.5cm}\textbf{for  }$r \gets 1$ to $R$ \textbf{do}
\STATE\hspace{1cm} $\hat{\bm{y}}\gets $\texttt{\ac{ceqln}\_evaluation}$(\mathcal{D},\ecdataset^r,\inecdataset^r)$  
\STATE\hspace{1cm} $L=L+\ac{sse}(\hat{\bm{y}},\bm{y})$
\STATE\hspace{0.5cm}\textbf{end }
\STATE$\hspace{0.5cm}\bm{\theta}_{k+1}=\bm{\theta}_{k}+\alpha_\theta\nabla_\theta L$
\STATE$\hspace{0.5cm}\bm{\Phi}_\theta\gets\bm{\theta}_{k+1}$
\STATE\textbf{end }
\ENSURE  $\bm{\theta}^{*}$
\end{algorithmic}
\end{algorithm}

The approach proposed to train the \ac{ceqln} is summarized in Alg.~\ref{alg:CEQLN_training}.
For the training process, we provide the training data sets $\mathcal{D}$, along with $R$ different adaptations of the same task defined in the datasets $\ecdataset^{r}$ and $\inecdataset^{r}$ where the index $r= 1,\ldots,R$. Each set of constraints indexed by $r$ regresses a  different  trajectory $\hat{\bm{y}}$, but shares the same quadratic function as defined in \eqref{eq:cqp}, which aims to minimize the distortion of the adapted trajectories regarding the training data set. The inclusion of multiple adaptation data sets aims to enhance the algorithm's generalization capabilities. The algorithm also requires the initial range values of the parameters $\bm{\theta}$ for the \ac{eqln}, denoted as $[\theta_a, \theta_b]$, such that for each parameter $\theta_i \in \bm{\theta}$ it holds that $\theta_a \leq \theta_i \leq \theta_b$. Additionally, the algorithm requires the number of initial evaluations of the \ac{ceqln}, denoted as $\beta$, and the number of epochs, where $\beta<\varepsilon$

The training algorithm consists of two main stages: initialization and training. 
\subsubsection*{\textbf{Initialization Stage}}
In this stage, the goal is to find an initial approximation of the parameters $\theta$ that can lead to faster convergence during training.
The algorithm evaluate every adaptation data set $\ecdataset^{r}$ or $\inecdataset^{r}$ or both (if applicable to the task), along with the complete training data set $\mathcal{D}$ using the \texttt{CEQLN.evaluation} function.   

\begin{algorithm}[t]
\caption{\ac{ceqln} Evaluation}\label{alg:CEQLN_evaluation}
\begin{algorithmic}[1]
\REQUIRE $\mathcal{D}=\left \{ \bm{t},\bm{y} \right \}$, $\ecdataset=\left \{ \ect,\ecy \right \}$, $\inecdataset=\left \{ \tineq,\yl,\yu \right \}$, $\lambda_w$
\STATEx\textbf{Block 1:} Forward pass on \ac{eqln}.\\
$\hspace{0.5cm}\phi_i(\bm{t})$, $\phi_i(\ect)$, $\phi_i(\tineq)$, $i=1,\ldots,M-1$ 
\STATEx\textbf{Block 2:} Concatenate $\phi_i$  with $\bm{1}\in\mathbb{R}^{\mathrm{N}}$,
$\oneseq\in\mathbb{R}^{\mathrm{P}}$, $\onesineq\in\mathbb{R}^{\mathrm{Q}}$.
\STATE $\bm{\Phi}=[\bm{1}^{\top},\phi_1(\bm{t})^{\top},\cdots,\phi_{M-1}(\bm{t})^{\top}]$ 
\STATE $\tilde{\bm{\Phi}}=[\oneseq^{\top},\phi_1(\ect)^{\top},\cdots,\phi_{M-1}(\ect)^{\top}]$
\STATE $\bar{\bm{\Phi}}=[\onesineq^{\top},\phi_1(\tineq),\cdots,\phi_{M-1}(\tineq)^{\top}]$
\STATEx\textbf{Block 3:} Build and solve the \ac{qp}.
\FOR {$d\gets 1$ to $D$}
\STATE $\bm{M}_d$, $\bm{c}_d$, $\dmeq_d$, $\dmineq_d$ 

\STATE$\bm{c}\gets[\bm{c}_1^{\top}, \cdots, \bm{c}_D^{\top}]^\top$
\STATE $\bm{P}_d\gets2\bm{M}^{\top}\bm{M}+2\lambda_w\bm{I}$
\STATE $\bm{q}_d\gets-2\bm{M}^{\top}\bm{c}$

\STATE $\bm{w}_d^{*}\gets$ \texttt{\ac{qp}\_solve}$(\bm{P}, \bm{q}, \dmeq, \ecy, \dmineq, \yl, \yu)$ 
\STATEx \hspace{0.5cm}\textbf{Block 4:} Calculate constrained regression trajectory.
\STATE$\hat{\bm{y}}_d\gets \bm{\Phi}\bm{w}_d^{*}$
\ENDFOR
\ENSURE  $\hat{\bm{y}}_d$
\end{algorithmic}
\end{algorithm}

For each evaluation, the parameters $\bm{\theta}$ are initialized using a uniform random distribution within the specified range $[{\theta}_a, {\theta}_b]$. The loss function $L$ is then calculated by comparing the regressed trajectory $\hat{\bm{y}}$ to the target trajectory $\bm{y}$. We perform $\beta$ evaluations in this stage, and each evaluation is stored in a list $\mathcal{G}$ as a tuple of $\bm{\theta}$ and $L$.

\subsubsection*{\textbf{Training Stage}}
In this stage we initialize the parameters $\bm{\theta}$ of the \ac{eqln} using the best parameters $\bm{\theta}$ obtained from $\mathcal{G}$ during the initialization stage. 
Afterwards, we evaluate again each data set  $\mathcal{D}$, $\ecdataset^{r}$, $\inecdataset^{r}$ and calculate the loss function $L$ for each set of constraints $r$ using the \texttt{\ac{ceqln}.evaluation} function. We then use gradient descent to backpropagate the error and update the parameters $\theta$. This is repeated  for the defined number of epochs. Through this iterative process, the design matrix $\bm{\Phi}$ is improved as the parameters of the \ac{eqln} are learned, resulting in a better fit between the regressed trajectory and the target trajectory $\bm{y}$. By performing this stage, the \ac{ceqln} algorithm ensures that the regressed trajectory $\hat{\bm{y}}$ satisfies the desired adaptations imposed by the constraints. Moreover, through the backpropagation process during training, the distortion between the regressed trajectory and the target trajectory $\bm{y}$ is minimized, leading to a closer fit between the two.



\subsection{\ac{ceqln} generalization}\label{subsec:Adaptation_without_retraining} 
As experimentally shown in Sec.~\ref{sec:validation}, the \ac{ceqln} demonstrates strong generalization capabilities by enabling the adaptation of trajectories to new conditions that were not included in the training phase. 
This capability arises from the fact that, during the training process, the spatial properties of the task are captured and encoded in the parameters of the \ac{bf} represented by $\bm{\theta}$. As a consequence, the trained \ac{eqln} acquires a learned understanding of the task's spatial properties. Then, to improve the generalization, the parameters of the \ac{eqln} $\bm{\theta}$ are fixed, and the \ac{qop} is solved again with the constraints corresponding to the new desired adaptation conditions. By solving the \ac{qop}, new values for the last layer $\bm{w}^{*}$ are obtained. These updated values capture the adaptations necessary to fulfill the new constraints. Thus, by reusing the trained \ac{eqln} and updating only the \ac{qop}, the \ac{ceqln} allows for efficient adaptation of the trajectory without the need for retraining the entire network.

\section{Validation}\label{sec:validation}

In this section, we validate \ac{ceqln} on a 2D dataset and compare its performance against $3$ baselines, namely \textit{\ac{cqp}}~\cite{CQP_Hector}, \textit{\ac{tpgmm}}~\cite{calinon2016tutorial}, and \textit{\ac{tpeqln}}~\cite{Villeda-2022-RAS}. This experiment is useful to understand \ac{ceqln} adaptation capabilities, in particular the possibility to adjust start and end of the trajectory through equality constraints. The parameters used for each approach are summarized in Tab.~\ref{table:Parameters_methods}.



\subsection{$2D$ letters} \label{section: 2D dataset}
\begin{figure*}[t]
    \centering
    \includegraphics[width=1.0\textwidth]{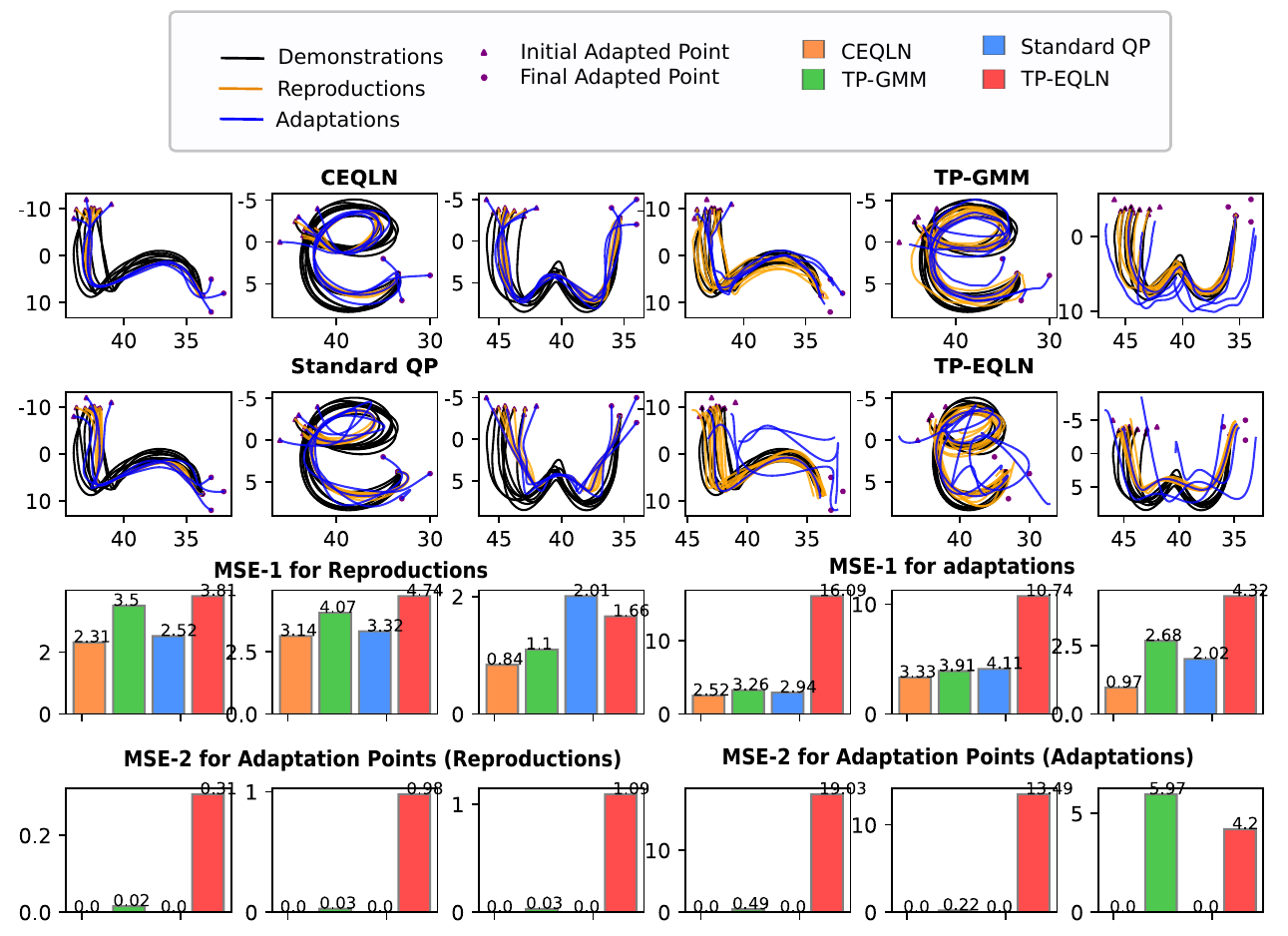}
    \caption{Results for the 2D data set. The first and second rows show the performance of each method for reproduction and adaptation cases.  Third and fourth rows show the $MSE_{\mathrm{shape}}$ and the $MSE_{\mathrm{const}}$ for reproduction and adaptation cases.} 
    \label{fig:2D_letter_exp}
\end{figure*}

\begin{figure*}[t]
    \centering
    \includegraphics[width=1.0\textwidth]{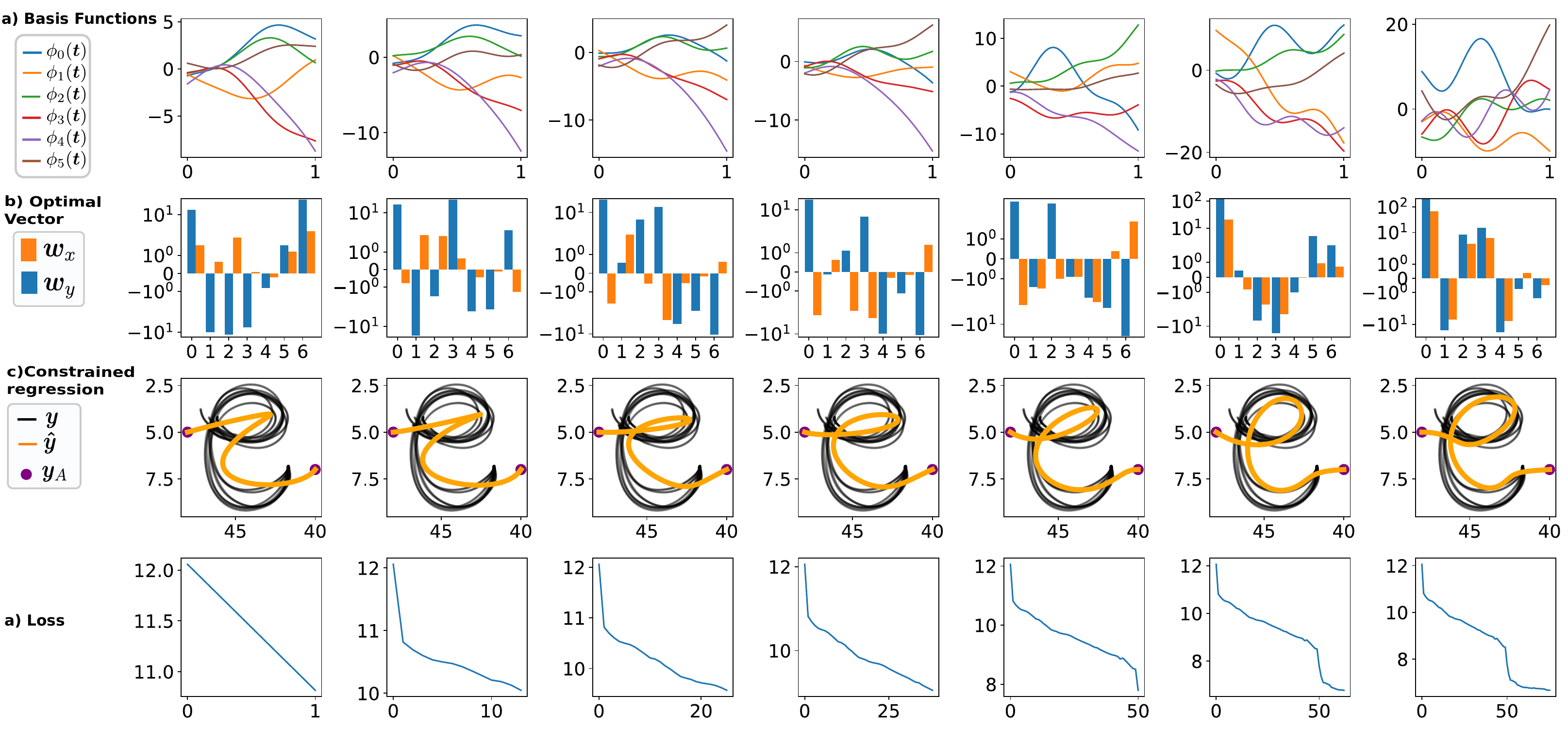}
    \caption{\ac{ceqln} training evolution. It is observed that the \ac{ceqln} can consistently satisfy the desired adaptations of the trajectory, even from the early stages of training (first column). As the training progresses, the \ac{eqln} model better fits the \ac{bf}, minimizing the distortion of the trajectories. In the second row, we can observe seven optimal vectors $\bm{w}$ for each dimension. However, there are only six \ac{bf} present. This is because the first element of the optimal vector corresponds to the multiplication with the ones vector of the design matrix. Therefore, it serves as a bias term, allowing the model to account for any constant offset in the trajectory.} 
    \label{fig:CEQLN_evolution}
\end{figure*}
The data set consist of $3$ letters from the English alphabet, namely \texttt{h}, \texttt{e}, and \texttt{w}, shown in Fig.~\ref{fig:2D_letter_exp}. The dataset is collected using kinesthetic teaching with a Franka Emika robot~\cite{haddadin2022franka}, as discussed in \cite{auddy2022continual}. 
Each letter is demonstrated $8$ times and each demonstration consists of $1000$ data points ($x$--$y$ positions of the robot's end-effector). For each demonstration, we generate a time vector containing $1000$ equally spaced times between $0$ and $1$. The time vectors $\bm{t}\in\mathbb{R}^{1\times8000}$ and the position trajectories $\bm{y}\in\mathbb{R}^{2\times8000}$ form the training data set is $\mathcal{D}
=\left \{ \bm{t},\bm{y} \right \}$. We consider $4$ different set of adaptations by changing initial and final points as shown in Tab.~\ref{table:Adaptation_points_2D}. Adaptations are denoted by $\ecy\in\mathbb{R}^{4}$. For adaptation $r=4$, the initial and final points are obtained by averaging the corresponding points from the demonstrations. The adaptations define the data set $\mathcal{\tilde{D}}^r
=\left \{ \ect,\ecy \right \}$. 

\ac{ceqln} and \ac{cqp} model these adaptations as equality constraints $\dmeq\bm{w} =\ecy$ , where $\dmeq\in\mathbb{R}^{4\times 2M}$ is the evaluation  of the design matrix $\bm{\Phi}(\ect)$ for $\ect=[0,1]$. More in detail, the matrix $\dmeq$ is defined as
\begin{equation*}
\dmeq=
\begin{bmatrix}
\bm{\Phi}(0) & \bm{0}_{1\times M} \\
\bm{0}_{1\times M} & \bm{\Phi}(0)\\
\bm{\Phi}(1) & \bm{0}_{1\times M} \\
\bm{0}_{1\times M} & \bm{\Phi}(1)
\end{bmatrix},
\end{equation*}
where $\bm{0}_{1\times M}\in\mathbb{R}^{1\times M}$ is a vector of zeros.


\begin{table}[t]
    \centering
\caption{Adaptation points for the 2D data set.}
    \label{table:Adaptation_points_2D}
    \resizebox{\columnwidth}{!}{ 
 	\begin{tabular}{cccccc}
 	\toprule
	\multicolumn{1}{l}{} & \multicolumn{1}{l}{} & \multicolumn{4}{c}{\sc{Adaptation} $\ecy$}     \\ 
    \cmidrule{3-6}
 	\sc{Letter} & \sc{Time} $\ect$ & $r=1$ & $r=2$ & $r=3$ & $r=4$ \\
 	\midrule
 	\multirow{2}{*}{h} & $0.0$ & $[44.0,-8.0]$ & $[41.0,-11.0]$ & $[43.0,-12.0]$ & $[42.7,-9.7]$ \\
 	 & $1.0$ & $[32.0,8.0]$ & $[33.0,-5.0]$ & $[33.0,12.0]$ & $[33.7,8.5]$ \\
 	  	\midrule
 	 \multirow{2}{*}{e} & $0.0$ & $[46.0,0.0]$ & $[42.0,-4.0]$ & $[44.0,-3.0]$ & $[43.1,-1.2]$ \\
 	 & $1.0$ & $[30.0,4.0]$ & $[35.0,2.0]$ & $[33.0,7.0]$ & $[33.4,3.8]$ \\
 	 \midrule
 	 \multirow{2}{*}{w} & $0.0$ & $[46.0,-5.0]$ & $[42.0,-4.0]$ & $[43.0,-3.0]$ & $[44.4,-3.5]$ \\
 	 & $1.0$ & $[36.0,-4.0]$ & $[34.0,-2.0]$ & $[34.0,-5.0]$ & $[35.3,-2.8]$ \\
	\bottomrule
\end{tabular}
}
\end{table}

\subsubsection{Performance evaluation} 
In Fig.~\ref{fig:2D_letter_exp}, the first and second rows illustrate the performance of each method for different letters.  For each of them, we show in yellow the reproduced trajectories---obtained considering the same initial and final point of the demonstrations---and in blue the adaptations---obtained considering the initial and final points in Tab.~\ref{table:Adaptation_points_2D}. 
To evaluate the accuracy of each method for both reproduction and adaptations, we consider two different MSE measures. The first $MSE_{\mathrm{shape}}$, shown in the third row of Fig.~\ref{fig:2D_letter_exp}, measures the distortion of the regressed trajectories and it is calculated as the MSE between the regressed trajectory and the training data points. Note that we compute the $MSE_{\mathrm{shape}}$ also for adaptations to measure the distrotion from the demonstrated data. The second $MSE_{\mathrm{const}}$, depicted in the fourth row of Fig.~\ref{fig:2D_letter_exp}, measures how effectively each method can fulfill the desired adaptations and it is calculated as the MSE between the adaptation point at the training points at $\ect$. Note that we compute the $MSE_{\mathrm{const}}$ also for reproductions to evaluate which approaches are capable of holding the initial and final points in the demonstrations.

Regarding the \ac{tpeqln}, it uses the initial/final point and time-step as task parameter. Its $MSE_{\mathrm{shape}}$ indicates poor performance, especially  for the adaptation points. This can be attributed to the variation in demonstration speeds. While the position-level characteristics of the demonstrations are quite similar, differences in speed parameterization can lead to variations in the timing of the trajectories. Consequently, two demonstrations with substantially different speeds may appear similar at the position level but exhibit temporal offsets when parameterized with respect to time. This temporal disparities between might affected to \ac{tpeqln} performance.

On the other hand, the \ac{tpgmm} method demonstrates better performance by preserving the shape of the trajectories in both reproduction and adaptation cases. However, it struggles to reach adaptation points in certain instances, as evident in the plots for the letter \textsl{w}, where the $MSE_{\mathrm{const}}$ is even larger than that of \ac{tpeqln}. 

In the case of the \ac{cqp} method, the overall shape of the trajectory is well-maintained but show some distortions, specially for the letter \textsl{w}. This may be attributed to either incorrect choice of the \ac{bf} or an insufficient number of them. On the other hand, for the adaptation points in both reproduction and adaptation cases $MSE_{\mathrm{const}}$ is zero, indicating that the constraints are perfectly met.  This is because the adaptation points are defined as equality constraints in \ac{qp}, and the method enforces the trajectories to converge towards these points.

Finally, \ac{ceqln} also achieves zero error for the $MSE_{\mathrm{const}}$, as adaptations are modeled as equality constraints in the \ac{qp}.  Furthermore, \ac{ceqln} exhibits the smallest $MSE_{\mathrm{shape}}$ for both reproduction and adaptation cases compared to the other methods. This is because, unlike the \ac{cqr} method, the \ac{eqln} determines the set the \ac{bf} via supervised learning.

To provide insights into the training process of \ac{ceqln}, Fig.~\ref{fig:CEQLN_evolution} displays seven key epochs for the letter \texttt{e} and adaptation $r=1$. The first row illustrates the improvement of the basis functions obtained from \ac{eqln}, while the second row depicts the optimal weights $\bm{w}^{*}=[\bm{w}_{x}^{*^\top},\bm{w}_{y}^{*^\top}]^\top$  that calculates the optimal linear combination of the basis functions for each dimension. The third row displays the regressed trajectory $\hat{\bm{y}}$, where the predicted trajectory increasingly fits the training data in each epoch. This is consistent with the decreasing loss displayed in the fourth row. The resulting trajectory $\hat{\bm{y}}$ always satisfies the adaptation points at $t=0$ and $t=1$ since they are enforced as equality constraints in \ac{ceqln}.

\subsubsection{Generalization capabilites}
\begin{figure}[t]
    \centering
    \includegraphics[width=1.0\columnwidth]{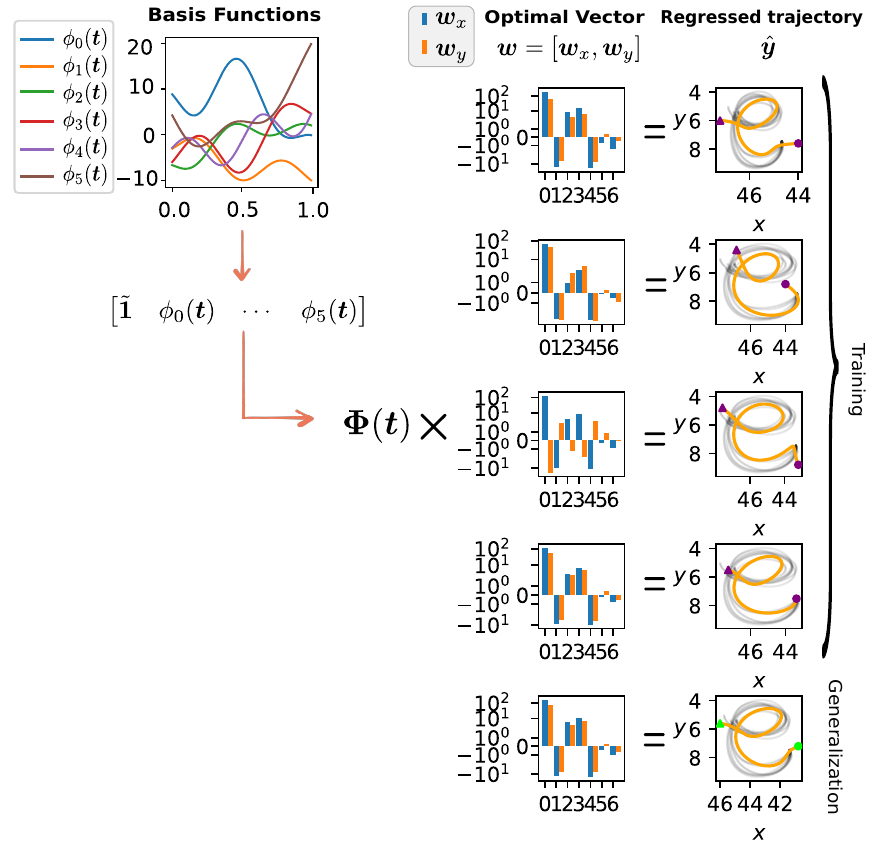}
    \caption{\ac{ceqln} generalization pipeline. Adapting the trajectory for different conditions involves solving the \ac{qop} to determine the optimal $\bm{w}$ that satisfies the imposed constraints. This approach guarantees minimal distortion of $\hat{\bm{y}}$, enabling the trajectory to be adapted efficiently for various conditions.}
    \label{fig:trajectory_adap_without_retraining}
\end{figure}

In Fig.~\ref{fig:trajectory_adap_without_retraining}, we show that \ac{ceqln} can generalize to new adaptations (initial and final points) without retraining. In particular, the \acp{bf} remain fixed, and for each adaptation $r$, the optimal weights $\bm{w}$ in the second column are re-computed, resulting in the constrained regressed trajectory displayed in the third column.
The first four rows show the behavior of \ac{ceqln} on the
$4$ adaptations in Rab.\ref{table:Adaptation_points_2D} (purple bullets). The approach clearly works well for data it has been trained on. In the last row, we demonstrate that the \ac{ceqln} can effectively generalize to new desired adaptation points given by $\ecy=[[45,-1],[32,3]]$ using the same set of \ac{bf}, i.e., without retraining. This demonstrates the model's robustness and versatility in handling new adaptation scenarios beyond the training data.

\subsubsection{Equation inspection} \label{ceqln_case_I}
Our approach is built upon the \ac{eqln} architecture, which allows us to obtain the analytical expression for the constrained regression model.  This provides valuable insights into the underlying equations. In particular, we can verify that the constraints $\dmeq\bm{w}= \ecy$ at $\ect=[0, 1]$ are fulfill. To demonstrate this, we evaluate $\bm{\Phi}$ with $\ect$, using the initial and final points $\ecy=[[45,-1],[32,3]]$ introduced in the generalization case shown in Fig.~\ref{fig:trajectory_adap_without_retraining}. In this case, the constraints write as
\begin{equation}\label{eq:Constraint_fulfilled_prube}
\tilde{\bm{\Phi}}\begin{bmatrix} {\bm{w}^*_x}^{\top}\\
{\bm{w}^*_{y}}^{\top}
\end{bmatrix}=\begin{bmatrix}
45 \\
-1 \\
32 \\
3 \\
\end{bmatrix},
\end{equation}
where $\bm{w}_x^{*}$ and $ \bm{w}_y^{*}$ are the optimal weights in $x$ and $y$ directions obtained by solving the \ac{qp}.
The design matrix $\bm{\Phi}$ is parameterized by the parameters $\bm{\theta}$ obtained through training the \ac{eqln}. 
By solving the matrix multiplication in~\eqref{eq:Constraint_fulfilled_prube}, as shown in Appendix~\ref{sec:Append_probe_constraints}, we can prove that the constraints are indeed fulfilled. This means that the predicted trajectory starts and ends at the desired adapted points defined by the constraints.

\section{Robot Experiments}\label{sec:real_esperiments}
In this section, we show $3$ experiments where a real Franka Emika Panda robot~\cite{haddadin2022franka} is asked to \textit{i) clean a surface}, \textit{ii) assemble mechanical parts}, and \textit{iii) place a bottle in a rack}. These experiments were designed to cover a range of scenarios and demonstrate the versatility of the \ac{ceqln} algorithm in handling various types of adaptations. More in detail, in the cleaning task, adaptations are based on the height of the surface to polish. This experiment demonstrates the effectiveness of \ac{ceqln} in adapting trajectories for complex manipulation tasks, where the desired adaptations are defined by specific task requirements. In the assembly task, adaptations are defined by the desired assembly point. This experiment highlights the ability of \ac{ceqln} to handle complex assembly scenarios, where the trajectory needs to be adapted to achieve precise positioning and alignment. In the bottle placing task, equality constraints are used to adapt the pick and place points of a bottle, while inequality constraints are employed to avoid obstacles. This experiment highlights the ability of \ac{ceqln} to handle multiple types of constraints simultaneously, enabling complex manipulation tasks in dynamic environments. In all the experiments, we present a comparison against $3$ baselines, namely \ac{tpgmm}, \ac{tpeqln}, and \ac{cqp}. The hyperparameters used in the $3$ experiments are reported in Tab.~\ref{table:Parameters_methods}. To enable differentiability in the operation $\bm{\Phi}\bm{w^{\top}}$, we decompose it as $w_{0_d}\cdot\bm{1}+[\phi_1(\bm{t}),\cdots,\phi_{\mathrm{M-1}}(\bm{t})]\bm{w}_{1:M_d}^{\top}$, allowing compatibility with \textit{TensorFlow}.
\textcolor{blue}{
}

\subsection{Cleaning task}
\begin{table}[t]
    \centering
\caption{Desired adaptations for cleaning task}
\label{table:Adaptation_cleanin_task_exp}
    \resizebox{\columnwidth}{!}{ 
 	\begin{tabular}{cccccc}
 	\toprule
	\multicolumn{1}{l}{} & \multicolumn{1}{l}{} & \multicolumn{4}{c}{{\sc{Adaptation}} $\ecy$ [m]}     \\ 
 	\cmidrule{3-6}
 	\sc{Type} & {\sc{Time}} $\ect$ [s] & $r=1$ & $r=2$ & $r=3$ & $r=4$ \\
 	\midrule
 	\sc{Initial point} & $\tilde{t} = 0.0$ & \multicolumn{4}{c}{$[0.46,0.0,0.09]^{\top}$} \\
 	\sc{Final point} & $\tilde{t} = 1.0$ & \multicolumn{4}{c}{$[0.46,0.0,0.09]^{\top}$} \\
 	 \sc{Surface contact} &  $0.2\leq \tilde{t} \leq 0.8$ & $z=0.16$&$z=0.21$&$z=0.26$&$z=0.30$ \\
	\bottomrule
\end{tabular}
}
\end{table}

\begin{figure*}[h]
    \centering
    \includegraphics[width=1.0\textwidth]{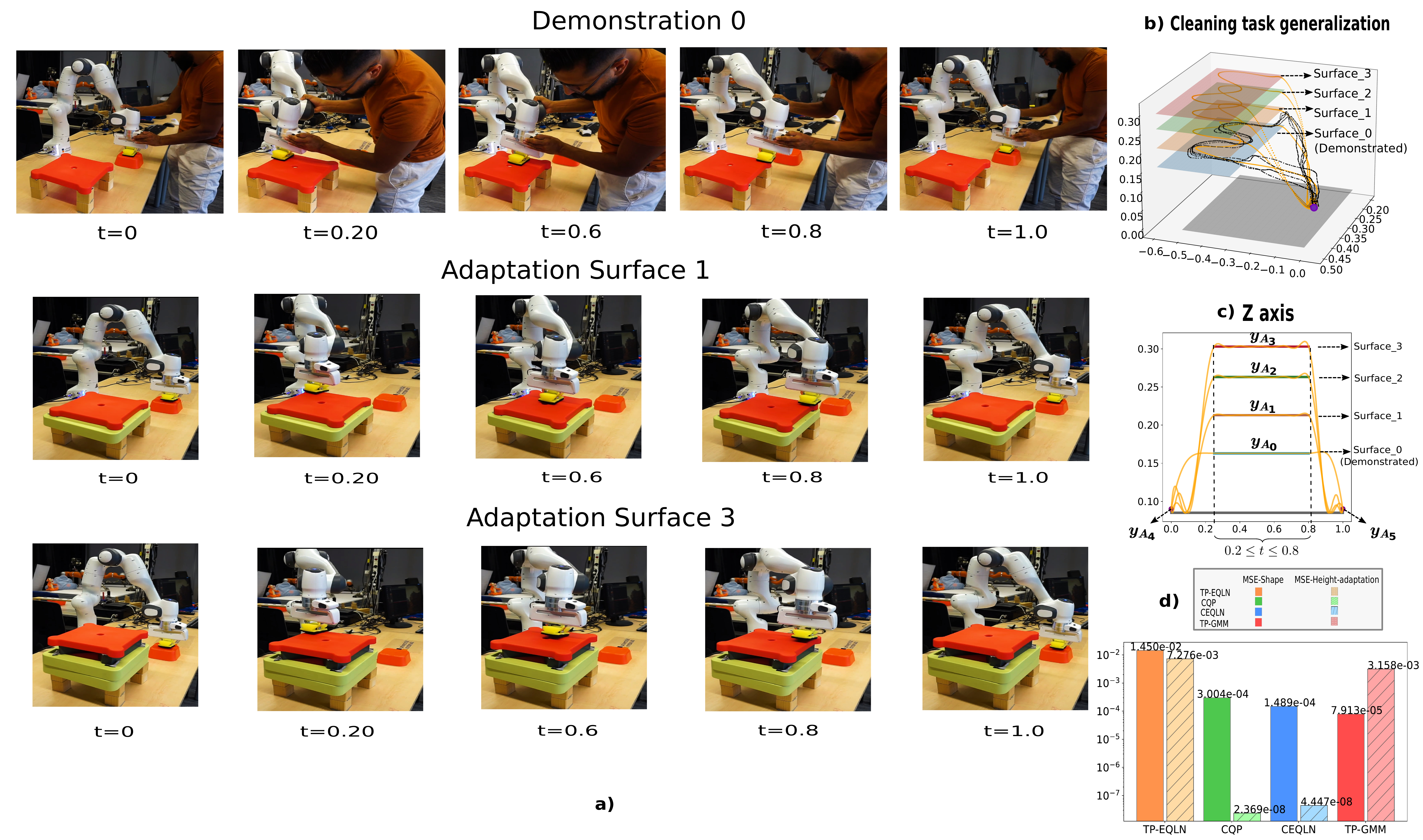}
    \caption{Cleaning task. Top-Left: Snapshots from one of the demonstrations used for training. Middle-Left and Bottom-Left: The robot performing the cleaning task on surfaces with different heights. Top-Right: Black trajectories representing the demonstrations used to train the \ac{ceqln} algorithm. Orange trajectories depict the adapted trajectories for surfaces with varying heights. Middle-Right: The Z-axis component of the trajectory demonstrating how the contact surface constraints for $r=1, \ldots, 4$ are fulfilled. Bottom-Right: MSE obtained for each method. Uniform color bars indicate the distortion of the obtained trajectory compared to the demonstrations, while light-lined bars measure the contact error between the surface and the obtained trajectory. .
}
    \label{fig:cleaning_task}
\end{figure*}

In this experiment, the robot has to clean surfaces placed at different heights. We collect $4$ demonstrations (Fig.\ref{fig:cleaning_task}a)) with $500$ $3$D positions each (black lines in the right-top plot of Fig.~\ref{fig:cleaning_task}) by placing the surface at a fixed height of $0.16\,$m (blue plane). Demonstrations form the training data set $\mathcal{D}
=\left \{ \bm{t},\bm{y} \right \}$ where $\bm{t}\in\mathbb{R}^{1\times2000}$ and $\bm{y}\in\mathbb{R}^{3\times2000}$. We consider the adaptations listed in Tab.~\ref{table:Adaptation_cleanin_task_exp}, forming the data set $\ecdataset
=\left \{ \ect,\ecy \right \}$. They are designed to: \textit{i)} start from the same initial point; \textit{ii)} perform the cleaning task at $3$ different heights (orange, green, and red planes in Fig.~\ref{fig:cleaning_task}); and \textit{iii)} end in the same final point. Note that initial and final points coincide in this task. Although we keep the initial and final points fixed in the $4$ adaptations, we still include them as constraints to ensure consistent pickup and release of the sponge.
%
%
The constraint on the height of the surface enforces the contact between the end-effector and the surface during the cleaning task. As this adaptation varies only in the height dimension ($z$-axis), we express its values as the height of the surface multiplied by a vector of ones $\bm{1}\in\mathbb{R}^{k}$. The constraint is active within the time interval $0.2 \leq \tilde{t} \leq 0.8\,$s, consisting of $k=100$ equally spaced time steps. The ranges of $\ect$ were obtained as an average directly from the demonstrations.

In \ac{cqp} and \ac{ceqln}, the adaptations in Tab.~\ref{table:Adaptation_cleanin_task_exp} are defined as equality constraints $\dmeq \bm{w}=\ecy$, where $\bm{w}\in \mathbb{R}^{3M}$ is obtained by solving the \ac{qp}, and $\ecy$ is a concatenated vector of each adaptation types (initial point, final point, and surface height). For example, in the case $r=1$, $\ecy=\left [2 \times \left [0.46,0.0,0.09 \right ]^{\top},0.16\cdot\bm{1} \right ]$.
The design matrix $\dmeq$ is defined as
\begin{equation*}
\dmeq=
\begin{bmatrix}
\bm{\Phi}(0) & \bm{0}_{1\times M} &\bm{0}_{1\times M}\\
\bm{0}_{1\times M}& \bm{\Phi}(0)   &\bm{0}_{1\times M}\\
\bm{0}_{1\times M} &\bm{0}_{1\times M}  &\bm{\Phi}(0) \\
\bm{\Phi}(1) & \bm{0}_{1\times M} &\bm{0}_{1\times M}\\
\bm{0}_{1\times M}& \bm{\Phi}(1)   &\bm{0}_{1\times M}\\
\bm{0}_{1\times M} & \bm{0}_{1\times M}  &\bm{\Phi}(1) \\
\bm{0}_{k\times M} & \bm{0}_{k\times M}   &\bm{\Phi}(\ect)
\end{bmatrix} \in\mathbb{R}^{(6+k)\times 3M},
\end{equation*}
where $\bm{0}_{r \times c}$ is a matrix of zeroes with $r$ rows and $c$ columns. Rows from $1$ to $3$ enforce the constraint on the initial $3$D position, while rows from $4$ to $6$ enforce the constraint on the final $3$D position. The block $\bm{\Phi}(\ect)\in\mathbb{R}^{k\times M}$, where $k=100$, represents the design matrix evaluated at $\ect = [0.2, \ldots, 0.8]^{\top}$, which relates to the surface height adaptation. In this case, only the $z$ dimension is considered, resulting in matrices of zeros $\bm{0}_{k\times M}$ for the $x$ and $y$ dimensions.


The results obtained with \ac{ceqln} on the cleaning task for different heights ($r=1,\ldots,4$)  are depicted in Fig.\ref{fig:cleaning_task}a) and \ref{fig:cleaning_task}b) (orange lines). The trajectories consistently start and end at the same points, indicating that constraints on the initial and the final position are fulfilled. Moreover, the robot effectively maintains contact with the surface, as imposed by the surface contact constraints. The impact of these constraints on the contact can be further observed in Fig.~\ref{fig:cleaning_task}c), which illustrates how the trajectories adhere to the surface throughout the time interval $0.2\leq \tilde{t} \leq 0.8\,$s.

Figure~\ref{fig:cleaning_task}d) shows the comparison of \ac{ceqln} against the $3$ baseline. 
In the figure, we display two bars for each method representing different MSE. $MSE_{\mathrm{shape}}$ represents the distortion of the trajectories obtained by each method. $MSE_{\mathrm{const}}$ measures the error between the desired surface height and the height of the robot trajectory during the time interval $0.2\leq \tilde{t} \leq 0.8\,$s ($k=100$ steps).
\ac{tpeqln} has the highest $MSE_{\mathrm{shape}}$ and $MSE_{\mathrm{const}}$. This probably depends on the way we collected the training data, i.e., considering only $1$ height ($0.16\,$m). Indeed, \ac{tpeqln} considers the height as a task-parameter to generalize the trajectory. Having demonstrated only one value for the task-parameter limits the ability of \ac{tpeqln} to properly encode the height variability. On the other hand, \ac{tpgmm} has a slightly smaller $MSE_{\mathrm{shape}}$, meaning it can accurately reproduce the shape of the demonstrations. However, the $MSE_{\mathrm{const}}$ is quite high meaning it fails to keep the desired height. This is because the frames used to adapt the trajectory fall outside the data distribution, and, as a consequence, the performance of \ac{tpgmm} is significantly affected in the extrapolation domain. Clearly, loosing contact with the surface cause the cleaning task to fail.
\ac{cqp} has the second smallest $MSE_{\mathrm{shape}}$ and also exhibits very small values for $MSE_{\mathrm{const}}$. 
Finally, the \ac{ceqln} algorithm yields the smallest MSE for both measures, indicating that our method minimizes trajectory distortion and maintains better contact with the adapted surface height during the cleaning motion. 

\subsection{Assembly task}
\begin{figure*}[h]
    \centering
    \includegraphics[width=1.0\textwidth]{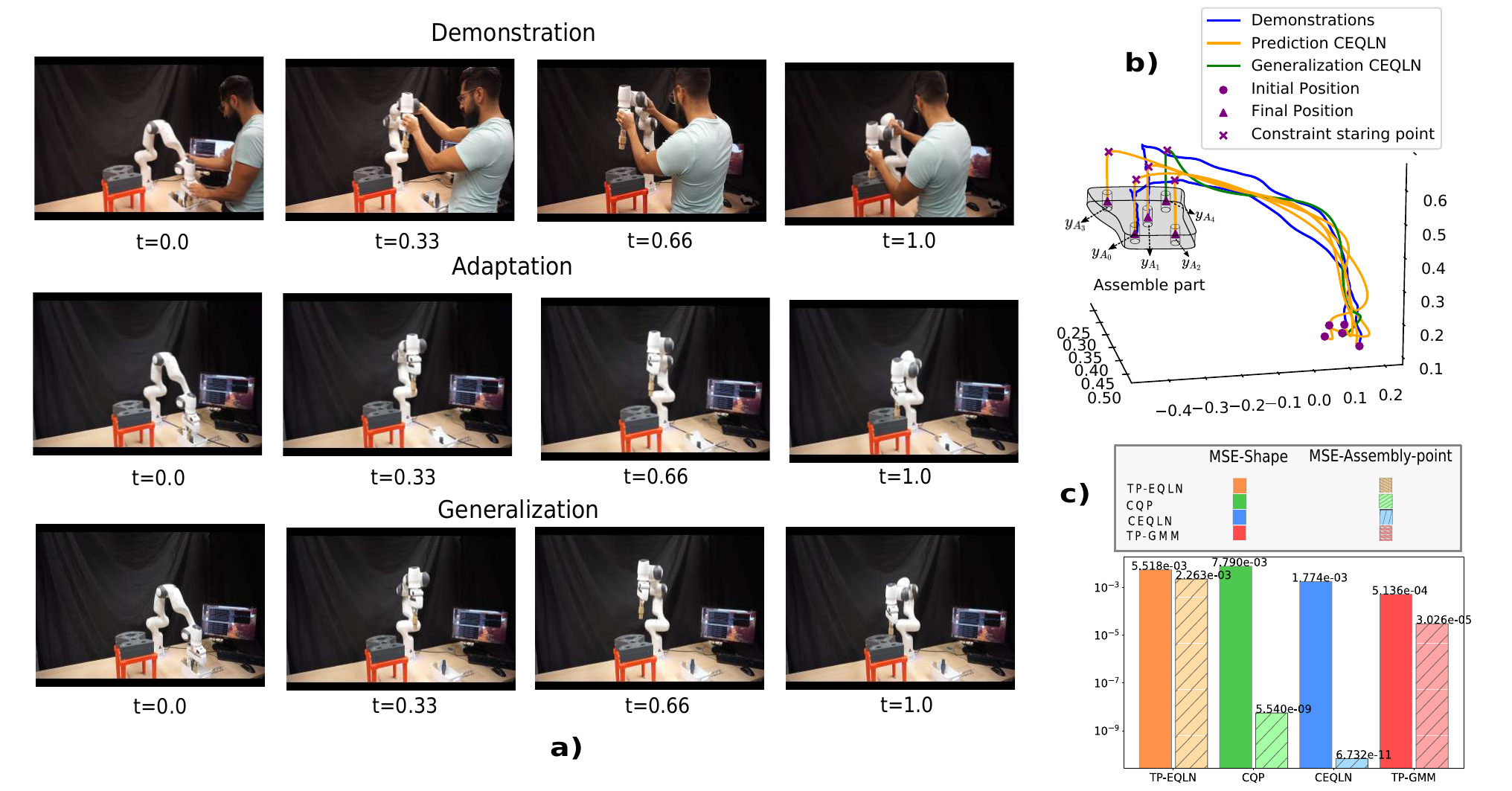}
    \caption{Assembly task: On first raw left  the demonstrations provided for training. On second-raw left the execution of one of the adapted trajectories for a different assemble goal. On third-raw left shows the generalization capabilities of the \ac{ceqln}, from which the goal was not used during the training process. Top-right shows the demonstrations, predictions and generalization trajectories obtained with \ac{ceqln}. Bottom-right, the MSE obtained for each method. Uniform color bars represent the distortion error of the obtained trajectory regarding the demonstrations and light-streaked bars represent the deviation of the last point of the trajectories regarding the assembly goal.  }
    \label{fig:assemble_task}
\end{figure*}
In this experiment, the robot has to perform a precision assembly task. We introduce variations in the initial and final (assembly) position to further increase the complexity of the task. We collect $2$ demonstrations with $500$ data points each (see Fig.~\ref{fig:assemble_task}a)), ending-up in $2$ different goals, to create the training dataset $\mathcal{D}
=\left \{ \bm{t},\bm{y} \right \}$ where $\bm{t}\in\mathbb{R}^{1\times1000}$ and $\bm{y}\in\mathbb{R}^{3\times1000}$. These demonstrations are depicted with blue lines in Fig.~\ref{fig:assemble_task}b). As shown in Tab.~\ref{table:Adaptation_assembly_task}, we consider $5$ scenarios ($r=1,\ldots,5$) by changing starting point, final point (assembly goal), and alignment assembly. The starting point is set at $\tilde{t}=0$, the assembly goal at $\tilde{t}=1$, and the alignment assembly within the interval $0.7 \leq \tilde{t} < 1.0$, which consists of $k=20$ equally spaced time steps. The alignment assembly constraint restricts the motion in the $x$ and $y$ axes while allowing freedom in the $z$ axis. Therefore, it ensures proper alignment of the part before placing it in the slot. Adaptations $r=1$ and $r=2$ are directly obtained from the demonstrations, while adaptations $r=3,4,5$ contain unforeseen points. Considered adaptations form the dataset  $\ecdataset^r =\left \{ \ect,\ecy \right \}$. 
Adaptations $r=1,\ldots,4$ are used to train \ac{ceqln} and learn the \ac{bf}, while adaptation $r=5$ is used to assess the generalization capabilities of \ac{ceqln}. 

\begin{table}[t]
    \centering
\caption{Desired adaptations for the Assembly task}
\label{table:Adaptation_assembly_task}
    \resizebox{\columnwidth}{!}{ 
 	\begin{tabular}{cccc}
 	\toprule
	\multicolumn{1}{l}{} & \multicolumn{3}{c}{{\sc{Time}} $\ect$ [s]}     \\ 
 	\cmidrule{2-4}
 	{\sc{Adaptation}} $\ecy$ [m] & $\tilde{t} = 0.0$ & $0.7\leq \tilde{t} \leq1.0$ & $\tilde{t} = 1.0$ \\
 	\midrule
  	$r=1$ & $[0.50,0.14,0.12]^{\top}$ & $x=0.33,y=-0.37$ & $[0.33,-0.37,0.42]^{\top}$\\
	$r=2$ & $[0.42,0.15,0.12]^{\top}$ & $x=0.39,y=-0.43$ & $[0.39,-0.43,0.42]^{\top}$\\
	$r=3$ & $[0.42,0.11,0.12]^{\top}$ & $x=0.40,y=-0.32$ & $[0.40,-0.32,0.42]^{\top}$\\
	$r=4$ & $[0.46,0.07,0.12]^{\top}$ & $x=0.24,y=-0.45$ & $[0.24,-0.45,0.42]^{\top}$\\
	$r=5$ & $[0.45,0.13,0.12]^{\top}$ & $x=0.25,y=-0.28$ & $[0.25,-0.28,0.42]^{\top}$\\
	\bottomrule
\end{tabular}
}
\end{table}



The design matrix is defined as
\begin{equation*}
\dmeq=
\begin{bmatrix}
\bm{\Phi}(0) & \bm{0}_{1\times M} &\bm{0}_{1\times M}\\
\bm{0}_{1\times M}& \bm{\Phi}(0)   &\bm{0}_{1\times M}\\
\bm{0}_{1\times M} &\bm{0}_{1\times M}  &\bm{\Phi}(0) \\
\bm{\Phi}(1) & \bm{0}_{1\times M} &\bm{0}_{1\times M}\\
\bm{0}_{1\times M}& \bm{\Phi}(1)   &\bm{0}_{1\times M}\\
\bm{0}_{1\times M} & \bm{0}_{1\times M}  &\bm{\Phi}(1) \\
\bm{\Phi}(\ect) & \bm{0}_{1\times M} &\bm{0}_{1\times M}\\
\bm{0}_{1\times M}& \bm{\Phi}(\ect)   &\bm{0}_{1\times M}\\

\end{bmatrix}\in\mathbb{R}^{(6+2k)\times 3M},
\end{equation*}
where rows from $1$ to $3$ enforce the constraint on the initial $3$D position, while rows from $4$ to $6$ enforce the constraint on the final $3$D position. The block $\bm{\Phi}(\ect)\in\mathbb{R}^{k\times M}$, where $k=20$, represents the design matrix evaluated at $0.7 \leq \tilde{t} < 1.0$, which relates to the alignment adaptation. In this case, $x$ and $y$ dimensions are considered, resulting in $2k=40$ extra rows in the design matrix. The desired adaptations are defined as equality constraints, express as $\dmeq \bm{w}=\ecy$, where  $\bm{w}\in \mathbb{R}^{3M}$. 

Results are shown in Fig.~\ref{fig:assemble_task}. In Fig.~\ref{fig:assemble_task}a), the first row displays snapshots from one of the demonstrations used for training, while the second and third rows show snapshots from the adapted trajectories for the adaptation $r=3$ and $r=5$ respectively. Figure \ref{fig:assemble_task}b) shows the results obtained from \ac{ceqln} for the constraints $r=1,\ldots,4$ contained in $\tilde{\mathcal{D}}$ (orange lines) and for the generalization $r=5$ (green line). As shown in the plots, all trajectories maintain a similar shape to the demonstrations, including the green trajectory, and converge to the desired assembly point.

We also compared the results obtained from \ac{ceqln} against $3$ baselines (see Fig.~\ref{fig:assemble_task}c)) in terms of $MSE_{\mathrm{shape}}$, which measures the distortion of the trajectory for each method, and $MSE_{\mathrm{const}}$, which measures the deviation between the final point of the predicted trajectory and the assembly goal. Trajectories generated by \ac{tpeqln} present high $MSE_{\mathrm{shape}}$ and $MSE_{\mathrm{const}}$, likely due to the limited number of demonstrations provided, leading to a low encoding level of the final point of the trajectory in the feature parameter space. Indeed, we only collect $2$ demonstrations in this experiment, and, therefore, also adaptations $r=3,4$ are also unknown to \ac{tpeqln}. 

Trajectories generated by \ac{cqp} converge well to the assembly goals, but present the highest distortion due to either an improper definition (number and type) of the \acp{bf} and/or suboptimal parameters. Trajectories generated by \ac{tpgmm} have the lowest distortion ($MSE_{\mathrm{shape}}$), but the deviation for the assembly goals ($MSE_{\mathrm{const}}$) is considerably high. This probably depends on the frames used to adapt the trajectories that are outside the data distribution. Finally, \ac{ceqln} generates trajectories with the second lowest distortion ($MSE_{\mathrm{shape}}$) and the lowest deviation error  regarding the assembly goal ($MSE_{\mathrm{const}}$), making it the most promising method for this task.

\subsection{Bottle pick and place with obstacle avoidance} 
\begin{figure*}[t]
    \centering
    \includegraphics[width=1.0\textwidth]{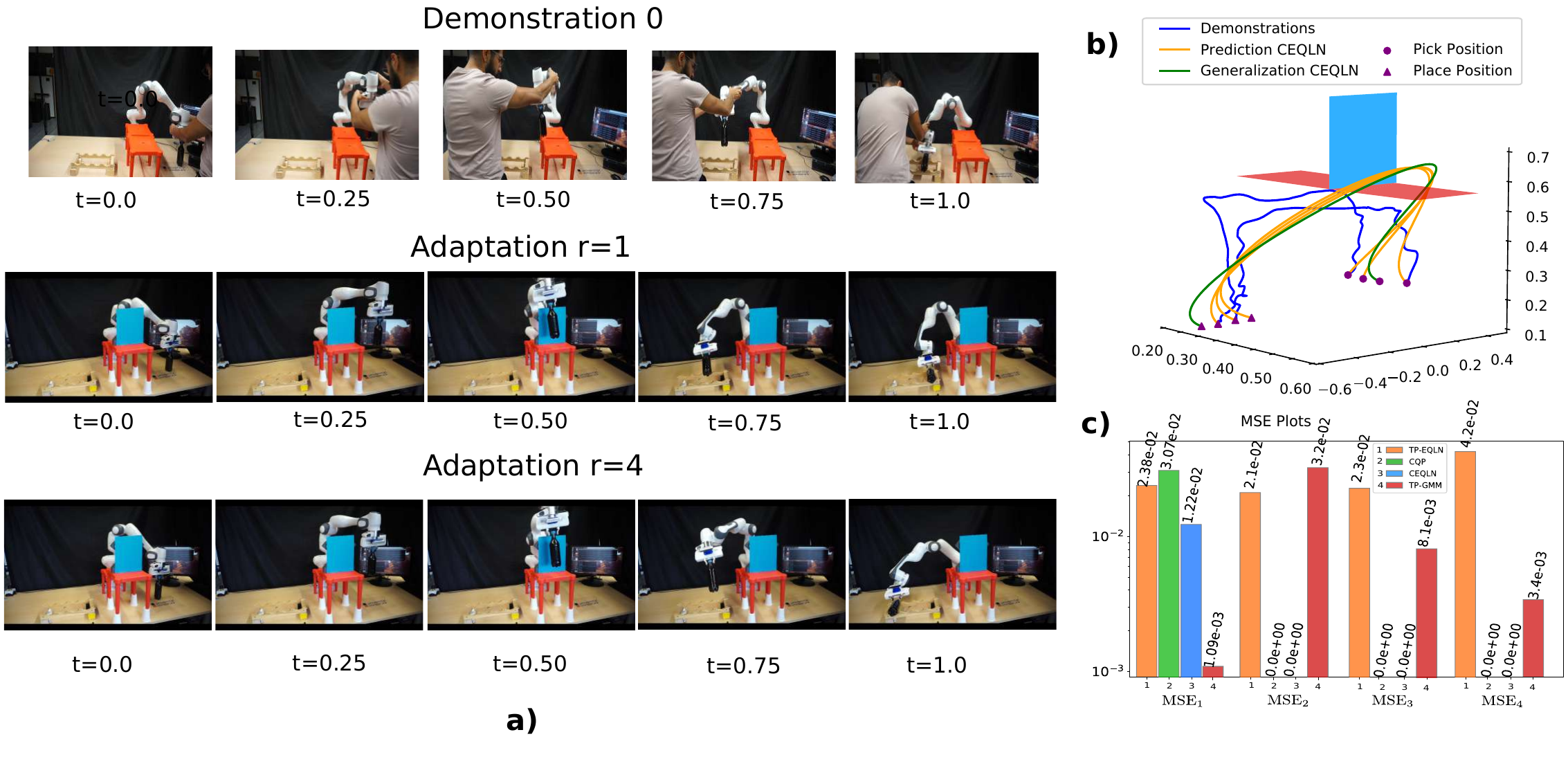}
    \caption{Bottle pick and place experiment. Left snapshots shows one demonstration used for training and some reproduced trajectories for new adaptations. Right-Top shows the adapted trajectories obtained from \ac{ceqln}. Left-Bottom a comparative MSE plots for the different desired adaptations between the different methods}
    \label{fig:Obstacle_avoidance_experiment}
\end{figure*}
In this experiment, we demonstrate the use of both equality and inequality constraints to accomplish the task of picking a bottle and placing it in a rack while avoiding a obstacles in the workspace. We provide $2$ demonstrations with $500$ points each (see Fig.~\ref{fig:Obstacle_avoidance_experiment}a)), representing different pick and place points and stored in the dataset $\mathcal{D}
=\left \{ \bm{t},\bm{y} \right \}$ where $\bm{t}\in\mathbb{R}^{1\times1000}$ and $\bm{y}\in\mathbb{R}^{2\times1000}$. We define $4$ sets of desired adaptations $r=1,\ldots 4$ (see Tab.~\ref{table:Stack_bottle_task}). Each set contains the desired pick (initial) and place (final) points, as well as the height ($z$ direction) and depth ($x$ direction) of the obstacle. The pick and place points for $r=2,3$ where obtained directly from the demonstrations. The place point for $r=1,4$ was obtained from the location of the slot $1$ (most left) and $4$ (most right) of the rack (see Fig.~\ref{fig:Obstacle_avoidance_experiment}). The obstacle avoidance behavior takes place within the time interval $0.3\leq \bar{t} \leq 0.65$, spanning $k=80$ equally spaced time steps. This interval was obtained from the demonstrations by identifying the time slot where the trajectory was going over the red table, kept fixed during the experiments. We use adaptations $r=1,2,3$ for training and adaptation $r=4$ for testing. 




\begin{table}[t]
    \centering
\caption{Desired adaptations for Stack Bottle task}
\label{table:Stack_bottle_task}
    \resizebox{\columnwidth}{!}{ 
 	\begin{tabular}{cccc}
 	\toprule
	{\sc{Adaptation}} & \multicolumn{3}{c}{{\sc{Time}} $\ect$, $\bar{\bm{t}}$ [s]}     \\ 
 	\cmidrule{2-4}
 	 $\ecy$, $\yl$ [m] & $\tilde{t} = 0.0$  & $0.3\leq \bar{t} \leq 0.65$ & $\tilde{t} =1.0$ \\
 	\midrule
 $r=1$	&$[0.40,0.40,0.21]^{\top}$ & $x\geq 0.55$, $z\geq 0.6$&  $[0.31,-0.34,0.13]^{\top}$ \\
 $r=2$	&$[0.26,0.35,0.21]^{\top}$ &  $x\geq 0.55$, $z\geq 0.6$ & $[0.31,-0.43,0.13]^{\top}$\\
 $r=3$ & $[0.31,0.33,0.21]^{\top}$ & $x\geq 0.55$, $z\geq 0.6$ & $[0.31,-0.53,0.13]^{\top}$ \\
 $r=4$ & $[0.35,0.34,0.21]^{\top}$ & $x\geq 0.55$, $z\geq 0.6$ & $[0.31,-0.62,0.13]^{\top}$\\
 	\bottomrule
\end{tabular}
}
\end{table}


\ac{ceqln} and \ac{cqp} model pick and place point adaptations as equality constrains. The design matrix for the equality constraints is defined as
\begin{equation*}
\dmeq=
\begin{bmatrix}
\bm{\Phi}(0) & \bm{0}_{1\times M} &\bm{0}_{1\times M}\\
\bm{0}_{1\times M}& \bm{\Phi}(0)   &\bm{0}_{1\times M}\\
\bm{0}_{1\times M} &\bm{0}_{1\times M}  &\bm{\Phi}(0) \\
\bm{\Phi}(1) & \bm{0}_{1\times M} &\bm{0}_{1\times M}\\
\bm{0}_{1\times M}& \bm{\Phi}(1)   &\bm{0}_{1\times M}\\
\bm{0}_{1\times M} & \bm{0}_{1\times M}  &\bm{\Phi}(1) \\
\end{bmatrix} \in \mathbb{R}^{6 \times 3M},
\end{equation*}
where rows from $1$ to $3$ enforce the constraint on the initial $3$D position, while rows from $4$ to $6$ enforce the constraint on the final $3$D position.
\ac{ceqln} and \ac{cqp} model collision avoidance adaptations as inequality constraints. The design matrix of the inequality constraints is defined as
\begin{equation*}
\dmineq=
\begin{bmatrix}
\bm{\Phi}(\ect) & \bm{0}_{k\times M} & \bm{0}_{k\times M}\\
\bm{0}_{k\times M} & \bm{0}_{k\times M} & \bm{\Phi}(\ect)  \\
\end{bmatrix} \in \mathbb{R}^{k \times 3M},
\end{equation*}
where the first $k=80$ rows relates to the object depth ($x$ direction), while the last $k$ rows relates the object height ($z$ direction). 

Trajectories generated by  \ac{ceqln} for each $r$ are shown in Fig.~\ref{fig:Obstacle_avoidance_experiment}b). The orange lines correspond to the constraints $r=1,2,3$ used for training, while the green line ($r=4$) shows a generalized trajectory. As observed in the plots, all the trajectories start and end at the desired pick and place locations and are successfully adapted to avoid the obstacles. Snapshots of the robot performing the task are shown in the second ($r=1$) and third ($r=4$) rows of Fig.~\ref{fig:Obstacle_avoidance_experiment}a). 

For the comparison with the baselines in Fig.~\ref{fig:Obstacle_avoidance_experiment}c), we use four MSEs:
\begin{equation*}
\resizebox{1.0\hsize}{!}{$
\begin{matrix}
 \textup{MSE}_1&=  &\sum_{r=0}^{r=4}\sum_{t\forall\in(\bm{t} \setminus \tineq)}\textup{MSE}(\bm{y},\hat{\bm{y}}^r) \\ 
 \textup{MSE}_2&=  &\sum_{r=0}^{r=4}\sum_{\bar{t}=0.3}^{0.65}\begin{cases}
    (0.55\cdot\bm{1}-\hat{\bm{y}}_x^r)^{2},& \text{if } (0.55\cdot\bm{1}-\hat{\bm{y}}_x^r)>0\\
    0,              & \text{otherwise}
\end{cases} \\ 
\textup{MSE}_3 &=  & \sum_{r=0}^{r=4}\sum_{\bar{t}=0.3}^{0.65}\begin{cases}
    (0.6\cdot\bm{1}-\hat{\bm{y}}_z^r)^{2},& \text{if } (0.6\cdot\bm{1}-\hat{\bm{y}}_z^r)>0\\
    0,              & \text{otherwise}
\end{cases}\\
\textup{MSE}_4 &=  &\sum_{t=1}
    (\bm{y}-\hat{\bm{y}}^r)^{2}
\end{matrix}$}
\end{equation*}
The MSEs evaluate different aspects of the trajectories. $\textup{MSE}_1$ measures the distortion in the trajectory segments before and after obstacle avoidance, which is defined by the time steps $t\in\{\bm{t} \setminus \tineq\}$. The index $r$ in $\hat{\bm{y}}$ defines the index regressed trajectory for each adaptation $r$. $\textup{MSE}_2$ measures how much the trajectory violates the constraint defined by the blue plane (object depth) during the obstacle avoidance, i.e., for $0.3\leq \bar{t}\leq 0.65$ given by $\tineq$. The subscript $x$ represents the $x$ component of $\hat{\bm{y}}$.  Similarity, $\textup{MSE}_3$ measures the violation of the constraint on the object height, where $\hat{\bm{y}}_z$ is the $z$ component of $\hat{\bm{y}}$. Finally, $\textup{MSE}_4$ measures the gap between the desired and reached final position.

Figure~\ref{fig:Obstacle_avoidance_experiment}c) shows that  \ac{tpeqln} has the highest MSEs values. The metrics indicates the method struggles to properly adapt the trajectory to avoid the new obstacles and faces difficulties to reach the goal. Similar to previous experiments, the reason of the low performance of this method is due to the low number of demonstrations, unable to adequately provide a variation of the task for each of the features of interest. These features are crucial in this method for encoding the physical properties of the task in the feature parameter space defined by time, goal, and scalar parameters for modifying the trajectory in the $x$ and $z$ dimensions.

\ac{tpgmm} preserves better the shape of the trajectory for the time steps the robot does not require obstacle avoidance. However, it has also difficulties to properly adapt the trajectory for the obstacle avoidance part as well as reaching the goal points. This is again attributed to the frames used to adapt the trajectories, which are defined outside the data distribution. 

For the  \ac{cqp} method, we observe excellent adaptation properties in avoiding obstacles and reaching the goals, as indicated by the values $\textup{MSE}_2=\textup{MSE}_3=\textup{MSE}_4=0$. However, this method shows the highest $\textup{MSE}_1$ value, suggesting that it introduces significant distortion in the trajectory. As for the previous experiments, this distortion may be due to suboptimal values of the \acp{bf} and/or to an improper definition (number and type) of the \acp{bf}.

Finally, our method demonstrates the second-lowest $\textup{MSE}_1$ value and adequately preserves the shape of the trajectory while also generating smooth motions, as shown in Fig.~\ref{fig:Obstacle_avoidance_experiment}c).  Regarding the other metrics, \ac{ceqln} achieves $\textup{MSE}_2=\textup{MSE}_3=\textup{MSE}_4=0$, indicating that it can generate trajectories that effectively avoid obstacles while precisely reaching the desired goals. This is possible maintaining a low number of \acp{bf} ($6$). 
because \ac{ceqln} exploits different types of \acp{bf} and optimally determines their parameters.

\section{Conclusions and Future Work}\label{sec:conclusion}

In conclusion, this paper introduces a novel supervised learning framework, called \ac{ceqln} for constrained regression problems within the domain of \ac{pbd}. The presented approach effectively addresses several significant challenges related to the adaptation of robotic trajectories.

Our method ensures that desired adaptations of demonstrated tasks are successfully achieved, even when these adaptations lie outside the data distribution. This capability  enhances the adaptability of robotic tasks to various scenarios while preserving the shape of trajectories and providing precise solutions for task adaptations.

Furthermore, our approach deal the lack of intuition about the structure of the \acp{bf} in constrained regression problems.  It also allows the modelling of desired adaptations as constraints in a \ac{qop}, which provides a close solution for the adaptations,  offering a systematic way to adapt trajectories in a accurate way in the regression space. 

The supervised learning stage considered in our method allows to encode the generalization of the task in the \acp{bf}. This allows the reusability of the fitted \acp{bf} to adapt the trajectory to new scenarios, making \ac{ceqln} a powerful tool for real-world robotic applications. On the other hand, employing \ac{eqln} as the method to determine the best \acp{bf} provides not only superior performance but also add the benefit of interpretable equations of the \acp{bf}, increasing the explainability and meaningful interpretation compared to standard \ac{nn} models.

The effectiveness of our approach is demonstrated in a set of robot trajectories that require adaptation due to changes in the environment. Overall, our approach shows promising results and shows superior performance regarding the other methods used for comparison.

Specifically, \ac{ceqln} manage better and with higher precision the adaptations compared to \ac{tpeqln}. While \ac{tpeqln} relies purely on the extrapolation capabilities of \ac{eqln}, it exhibits a growing error gap when queries extend beyond the data distribution. In contrast, \ac{ceqln} models adaptations as constraints, enabling it to provide close solutions within the regression space, even for queries outside the data distribution. 

Furthermore, \ac{ceqln} offers superior adaptability compared to \ac{tpgmm}, which relies on a probabilistic distribution conditioned by the data distribution. The ability to adapt trajectories with high precision beyond the data distribution is a distinctive strength of \ac{ceqln}.

Additionally, \ac{ceqln} minimizes trajectory distortion more effectively than \ac{cbat}. Furthermore, it allows to encode task generalization directly into the fitted \acp{bf} allowing to reuse them for new adaptation requirements. This is capability is not supported by \ac{cbat}.

In summary, our approach not only demonstrates superior adaptability and precision but also provides enhanced generalization capabilities and reusability performance, making it a valuable method for several robotic applications subject to continuous adaptations due to changes in the environment. This addresses the challenge of minimizing the number of required demonstrations  for the adaptation and significantly reduces time and effort.

As future work, we aim to expand the input parameters of \ac{ceqln}. Currently, it uses time as an input parameter. However, we intend to extend this to incorporate other parameters, such as task-parameters  similar to what \ac{tpeqln} employs. This would allow us to encode additional task properties into the \acp{bf}, potentially enhancing the model's adaptability and performance across a broader range of scenarios.

Another avenue for future research involves to explore the full potential of \ac{ceqln} and its constrained regression capabilities across in various domains, including physics-informed machine learning. Particularly, we are interested in applying it to dynamic systems identification under constraints, which holds significance in control systems, as these systems must satisfy stability conditions typically modeled as constraints of the Lyapunov function.

Additionally, we plan to conduct more in-depth analyses regarding the explainability of the fitted \acp{bf}. Understanding how these \acp{bf} capture and interpret the underlying data is crucial for gaining insights into the model.


\appendices

\section{Constrained Equation Inspection}
\label{sec:Append_probe_constraints}
The \ac{eqln} architecture used  in Sec.~\ref{section: 2D dataset} to reproduce the \texttt{e} letter consists of one hidden layer with $11$ elemental functions. The training parameters fitted and the elemental functions are the following
\begin{equation*}
\begin{array}{rclcl}
f_0&=& (-6.56\,t+0.5)\times(1.85\,t-0.97),\\ 
f_1&=&(6.676\,t-0.328),\\
f_2&=&\sin(-2.426\,t+1.163),\\
f_3&=&\cos(1.489\,t+0.623),\\
f_4&=&\sigma(-2.664\,t-1.417),\\ 
f_5&=& \text{sech}(-5.98\,t-0.594),\\
f_6&=&(-5.921\,t-1.137),\\
f_7&=&\sin(11.969\,t+0.053),\\
f_8&=&\cos(10.269\,t+1.097),\\
f_9&=&\sigma(4.697\,t-1.433),\\ 
f_{10}&=& \text{sech}(12.109\,t-1.524).\\
\end{array}
\end{equation*}

The output of the first hidden layer $\bm{z}^{(1)}$ is composed by the set of elemental functions as following
\begin{equation*}
\begin{array}{lllllll}
\bm{z}^{(1)}=&[f_0&f_1&f_2&f_3&f_4&\cdots\\
    &f_5&f_6&f_7&f_8&f_9&f_{10}]
\end{array}.
\end{equation*}
The training parameters of the last layer $\bm{\theta}^{(2)}$ have the following values
\begin{equation*}
\resizebox{1.0\hsize}{!}{$
\bm{\theta}^{(2)} = \begin{bmatrix}
 \bm{b}^{(2)^\top} \\
 \bm{W}^{(2)^\top}
\end{bmatrix}=\begin{bmatrix}
  2.909& -0.641&  1.133&-1.561& -1.078& -3.199\\
  1.385&  0.277&  1.094& 0.218&  0.374& -1.983\\
  4.346&  1.256& -1.821& -1.77& -6.415& -0.973\\
  3.68 & -2.506& -2.64 & 1.527&  4.636&  0.41\\
  1.356& -0.751&  0.164& -1.87 & -2.28 & -1.477\\
  1.314& -1.536& -2.699&  0.31 &  0.107&  1.294\\
 -1.464&  2.4  &  0.223&-0.159&  0.661& -2.191\\
 -1.401&  0.404& -1.525& 0.248&  3.883& -1.493\\
  4.907& -2.931&  1.332& -4.698&  1.179&  1.695\\
  0.561&  0.549& -1.002& -1.952& -0.081& -1.575\\
  -1.2 & -0.28 & -0.298&-1.084&  1.056&  0.293\end{bmatrix}.$
  }
\end{equation*}
By multiplying the output of the first layer $\bm{z}^{(1)}$ by the parameters of the second layer $\bm{\theta}^{(2)}$, and summing up the scalar bias of the last layer, the \ac{bf} can be calculated as
\begin{equation*}
\begin{array}{l}
\phi_0=z^{(1)}\bm{W}^{(2)^\top}_{1}-0.762,\\
\phi_1=z^{(1)}\bm{W}^{(2)^\top}_{2}+2.548,\\
\phi_2=z^{(1)}\bm{W}^{(2)^\top}_{3}-0.672,\\
\phi_3=z^{(1)}\bm{W}^{(2)^\top}_{4}-3.331,\\
\phi_4=z^{(1)}\bm{W}^{(2)^\top}_{5}-0.966,\\
\phi_5=z^{(1)}\bm{W}^{(2)^\top}_{6}-0.642.\\
\end{array}
\end{equation*}
where $\bm{W}^{(2)^\top}_{i}$ is the $i$-th column of $\bm{W}^{(2)^\top}$. From these, the design matrix $\mathbf{\Phi}(t)$ is defined as  
\begin{equation*}
\mathbf{\Phi}(t)=
\begin{bmatrix}
 1&\phi_0 &\phi_1&\phi_2&\phi_3&\phi_4&\phi_5
\end{bmatrix}.
\end{equation*}
Finally, the computed optimal weights are
\begin{equation*}
\resizebox{1.0\hsize}{!}{$
\begin{array}{lll}
\mathbf{w^{*}}&=&[w^{*}_x,w^{*}_y]^{\top}\\
    & =&\begin{bmatrix}
 158.07& -12.74&   5.66&  10.67&  -15.2&  -0.46&  -1.03 \\ 
 71.48&  -6.91&    3.53&   6.44&  -7.65&   0.29&  -0.66 
\end{bmatrix}^{\top}
\end{array}.$}
\end{equation*}

\begin{table*}[t]
    \centering
 \caption{Hyperparameters used for each method and for each experiment}
\label{table:Parameters_methods}
    \resizebox{\textwidth}{!}{ 
 	\begin{tabular}{ccccc}
 	\toprule
	\multicolumn{5}{c}{\sc{CEQLN}} \\ 
 	\midrule
 	 \sc{Parameters}  & \sc{2D Letters} & \sc{Cleaning} & \sc{Assembly} & \sc{Pick \& Place} \\
 	\midrule
 	\sc{Num. Layers} & $1$&$2$&3&$1$ \\
  \sc{Epochs} & \phantom{0}25\,000& \phantom{0}2\,000& 4000& 4000\\
 \sc{Func. per layer} & $2\times\{I, \sin, \cos, \sigma,  \text{sech},\times\}$&$3\times\{I, \sin, \cos, \sigma,  \text{sech},\times\}$&$2\times\{I, \sin, \cos, \sigma,  \text{sech},\times\}$&$2\times\{I, \sin, \cos, \sigma,  \text{sech},\times\}$\\
 $M-1$ & 7 &19 &10&6 \\
 $l_r$ & $0.01$&$0.01$&$1e^{-3}$&$1e^{-3}$ \\
 $\lambda_w$ & $0.01$&$0.01$&$0.01$&$0.001$\\
  $\beta$ & $10$&$25$&$10$&$10$\\
  \multirow{2}{*}{$[\theta_a,\theta_b]^{ i }$} & $\left \{ [-10,10], [-1,1]\right \}^{0}$;&$\left \{ [-5,5], [-1,1]\right \}^{0}$;$ \left \{ [-2,2], [-1,1]\right \}^{1}$&$\left \{ [-2,2], [-1,1]\right \}^{0}$;$ \left \{ [-2,2], [-1,1]\right \}^{1}$&$\left \{ [-3,3], [-3,3]\right \}^{0}$;$ \left \{ [-3,3], [-1,1]\right \}^{1}$\\
 &$\left \{ [-1,1], [-1,1]\right \}^{1}$&$\left \{ [-1,1], [-1,1]\right \}^{2}$&$\left \{ [-2,2], [-1,1]\right \}^{2}$;$ \left \{ [-2,2], [-1,1]\right \}^{3}$&\\
 	\bottomrule
 	\toprule
	\multicolumn{5}{c}{\sc{CQP}} \\ 
 	\midrule
 	 \sc{Parameters}  & \sc{2D Letters} & \sc{Cleaning} & \sc{Assembly} & \sc{Pick \& Place} \\
 	\midrule
 \multirow{2}{*}{\sc{Set of  \ac{bf} }}&$\{1, t, t^{2}, \sin{kt},  \cos{kt}\}$&$\{1, t, t^{2}, \sin{kt},  \cos{kt}\}$&$\{1, t, t^{2}, \sin{kt},  \cos{kt}\}$&$\{1, t, t^{2}, \sin{kt},  \cos{kt}\}$\\
   &with $k=\{10,20\}$&with $k=\{0.01,0.1,1,5,10,15,25\}$&with $k=\{0.1,1,5,10\}$&with $k=\{0.1,1\}$\\
   	$M-1$& $7$&19&10&$6$ \\
$\lambda_w$& $0.01$&$0.01$&$0.01$&$0.01$\\
 	\bottomrule
 	\toprule
	\multicolumn{5}{c}{\sc{TP-EQLN}} \\ 
 	\midrule
 	 \sc{Parameters}  & \sc{2D Letters} & \sc{Cleaning} & \sc{Assembly} & \sc{Pick \& Place} \\
 	\midrule
 	\sc{Num. Hidden Layers} & $2$&2 &2 &2 \\
\sc{Epochs} & \phantom{0}5\,000& \phantom{0}5\,000& \phantom{0}5\,000 & \phantom{0}5\,000\\
\sc{Num. Features} & $5$& $2$& $4$& $6$\\
\sc{Batch Size} & 500& 500& 500& 500\\
\sc{Func. per layer} & $2\times\{I, \times, \sin, \cos, \sigma, \mathrm{sech},\}$&$2\times\{I, \sin, \cos, \sigma,  \text{sech},\times\}$&$2\times\{I, \sin, \cos, \sigma,  \text{sech},\times\}$&$2\times\{I, \sin, \cos, \sigma,  \text{sech},\times\}$\\
$l_r$ & $0.001$&$0.001$ &$0.001$&$0.001$ \\
$\lambda_\theta$&$1e^{-6}e$&$1e^{-6}e$&$1e^{-6}e$&$1e^{-6}e$\\
 $[\theta_a,\theta_b]^{\left \{ i \right \}}$& $0.1*\left \{ [-\overline{\bm{y}_{D}},\overline{\bm{y}_{D}}], [-\overline{\bm{y}_{D}},\overline{\bm{y}_{D}}]\right \}^{\left \{  0,1,2\right \}}$&$0.1*\left \{ [-\overline{\bm{y}_{D}},\overline{\bm{y}_{D}}], [-\overline{\bm{y}_{D}},\overline{\bm{y}_{D}}]\right \}^{\left \{  0,1,2\right \}}$&$0.1*\left \{ [-\overline{\bm{y}_{D}},\overline{\bm{y}_{D}}], [-\overline{\bm{y}_{D}},\overline{\bm{y}_{D}}]\right \}^{\left \{  0,1,2\right \}}$&$0.1*\left \{ [-\overline{\bm{y}_{D}},\overline{\bm{y}_{D}}], [-\overline{\bm{y}_{D}},\overline{\bm{y}_{D}}]\right \}^{\left \{  0,1,2\right \}}$\\
 	\bottomrule
 	\toprule
	\multicolumn{5}{c}{\sc{TP-GMM}} \\ 
 	\midrule
 	 \sc{Parameters}  & \sc{2D Letters} & \sc{Cleaning} & \sc{Assembly} & \sc{Pick \& Place} \\
 	\midrule
 	\sc{Num. Frames} & $4$&$3$&$3$&$6$ \\
 \multirow{6}{*}{\sc{Frames position} }&$F_0=\bm{y}_{D_{\{t=0,i=7\}}}$ &$F_0=\bm{y}_{D_{\{t=0,i=0\}}}$ &$F_0=\bm{y}_{D_{\{t=0,i=0\}}}$ &$F_0=\bm{y}_{D_{\{t=0,i=1\}}}$ \\
   &$F_1=\bm{y}_{D_{\{t=0.3,i=7\}}}$ &$F_1=\bm{y}_{D_{\{t=1,i=0\}}}$&$F_1=\bm{y}_{D_{\{t=0.5,i=0\}}}$;&$F_1=[\bm{y}_{D_{\left \{t=0.3,i\right \}_x}},\bm{y}_{D_{\left \{t=0.3,i\right \}_y}},0.6]$\\
   &$F_2=\bm{y}_{D_{\{t=0.7,i=7\}}}$&$F_2=\bm{y}_{D_{\{t=0.3,i=0\}}}$&$F_2=\bm{y}_{D_{\{t=1,i=0\}}}$&$F_2=[\bm{y}_{D_{\left \{t=0.6,i\right \}_x}},\bm{y}_{D_{\left \{t=0.6,i\right \}_y}},0.6]$\\
   &$F_3=\bm{y}_{D_{\{t=1,i=7\}}}$& & &$F_3=[0.55,\bm{y}_{D_{\left \{t=0.3,i\right \}_y}},\bm{y}_{D_{\left \{t=0.3,i\right \}_z}}]$\\
   & & & &$F_4=[0.55,\bm{y}_{D_{\left \{t=0.6,i\right \}_y}},\bm{y}_{D_{\left \{t=0.6,i\right \}_z}}]$\\
   & & & &$F_5=\bm{y}_{D_{\{t=1,i\}}}$\\
\sc{Num. Gaussians} & 20& 30& 30& 30\\
$\lambda_\sigma$ & $4e^{-10}$& $1e^{-4}$& $1e^{-4}$& $1e^{-4}$\\
\bottomrule
\end{tabular}
}
\end{table*}

\section{Hyperparameters selection}
\label{sec:Append_parameters}
The parameters used for the experiments in Sec.~\ref{sec:validation} and Sec.~\ref{sec:real_esperiments} are listed in Tab.~\ref{table:Parameters_methods}.
For our approach (\ac{ceqln}), we parameterize trajectories based on time, where the input of the \ac{eqln} is the time step $t$, and the output is determined by the number of basis functions ($M-1$). The parameter $\lambda_w$ is the regularization constant used to solve the \ac{qp} in both methods \ac{ceqln} and \ac{cqp}. The $l_r$ term is the learning rate value used for training the \ac{ceqln} and \ac{tpeqln}. $\lambda_\theta$ is the regularization constant in the LASSO cost function of the \ac{tpeqln} method. $\lambda_\sigma$ is the regularization constant used in the \ac{tpgmm} method. Finally,  $[\theta_a,\theta_b]^{\left \{ i \right \}}$ is the range of the uniform random distribution used to initialize the parameters of each layer of the \ac{eqln} in the \ac{ceqln} and \ac{tpeqln} methods.

\section*{Acknowledgements}
This research has received funding from the European Union under the NextGenerationEU project iNest (ECS 00000043), the Horizon 2020 research and innovation programme (grant agreement no. 731761, IMAGINE), and from the Austrian Research Foundation (Euregio IPN 86-N30, OLIVER). 


%


\bibliographystyle{IEEEtran}

\begin{IEEEbiography}[{\includegraphics[width=1in,height=1.25in,clip,keepaspectratio]{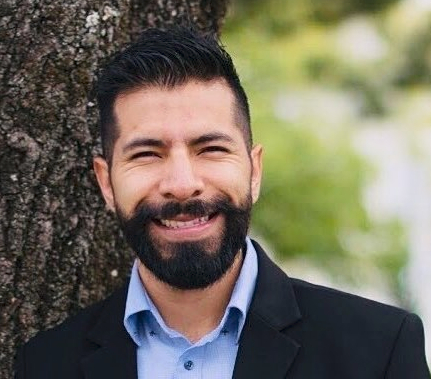}}]{Héctor Pérez Villeda}
received the M.Sc. degree in Robotics and Advanced Manufacturing at CINVESTAV, Mexico in 2015. In the same year he co-founded \textbf{Introid Inc.} , a company focused on A.I. and computer vision. In parallel, he continued collaborating with researchers from CINVESTAV, where he authored and co-authored different conference and journal papers focused on multi-robot coordination systems. In 2019 he was a research engineer at the Institute of Plasma and Nuclear Fusion (IPFN) in Lisboa, Portugal.

He is currently a Ph.D. candidate at the University of Innsbruck, Austria, and a member of the Intelligent and Interactive Systems group in the computer science department. His research is focused on Imitation learning, Programming by demonstrations, Machine learning for robot tasks generalization and optimization.
\end{IEEEbiography}

\begin{IEEEbiography}[{\includegraphics[width=1in,height=1.25in,clip,keepaspectratio]{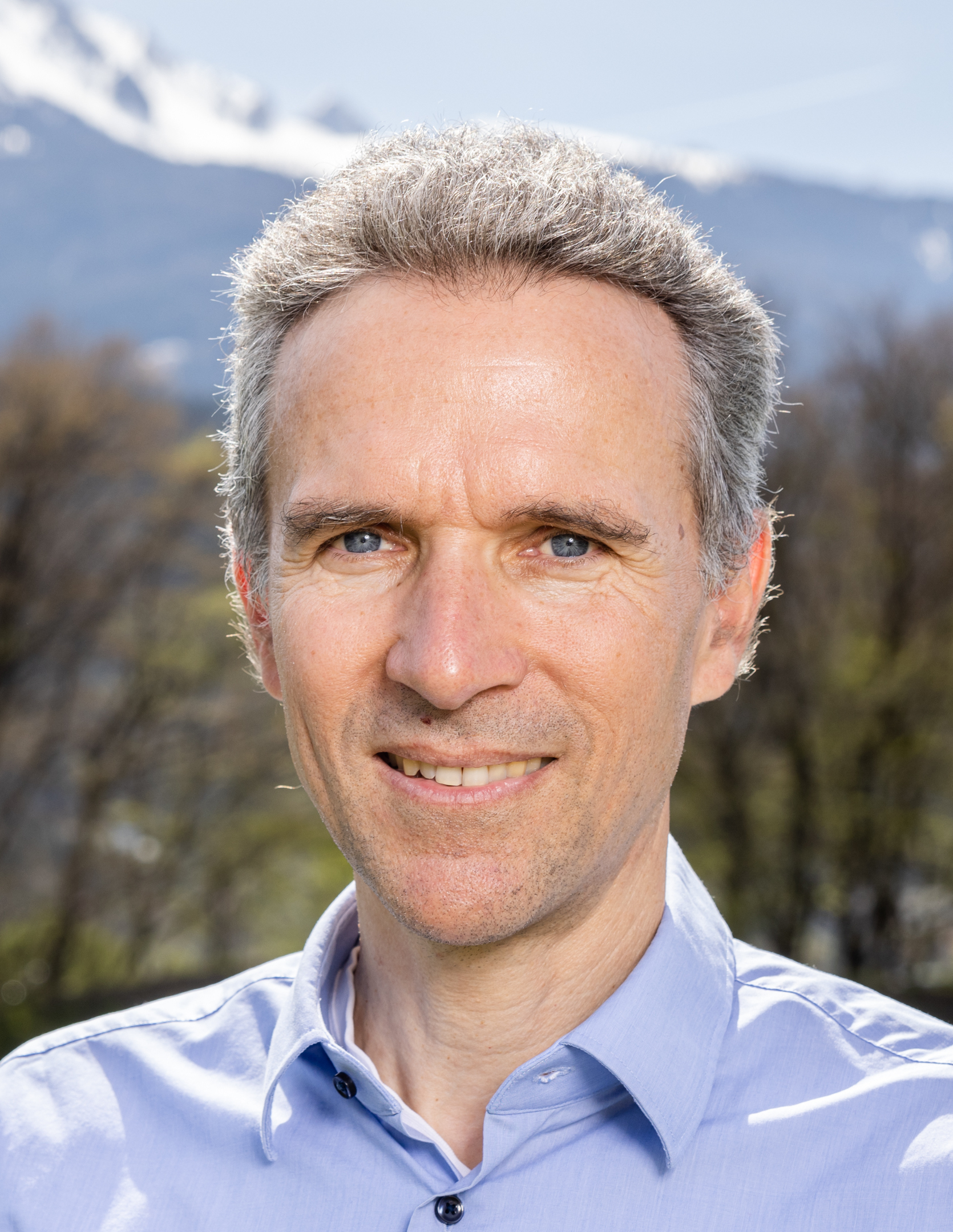}}]{Justus Piater}
is a professor of computer science at the University of Innsbruck, Austria, where he leads the Intelligent and Interactive Systems group. He holds a M.Sc. degree from the University of Magdeburg, Germany, and M.Sc. and Ph.D. degrees from the University of Massachusetts Amherst, USA, all in computer science. Before joining the University of Innsbruck in 2010, he was a visiting researcher at the Max Planck Institute for Biological Cybernetics in Tübingen, Germany, a professor of computer science at the University of Liège, Belgium, and a Marie-Curie research fellow at GRAVIR-IMAG, INRIA Rhône-Alpes, France. His research interests focus on learning and inference in sensorimotor systems. He has published more than 200 papers in international journals and conferences, several of which have received best-paper awards. Currently he serves as the founding director of the interdisciplinary Digital Science Center of the University of Innsbruck.
\end{IEEEbiography}

\begin{IEEEbiography}[{\includegraphics[width=1in,height=1.25in,clip,keepaspectratio]{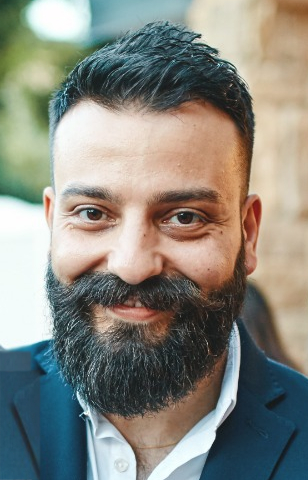}}]{Matteo Saveriano}
received his B.Sc. and M.Sc. degree in automatic control engineering from University of Naples, Italy, in 2008 and 2011, respectively. He received is Ph.D. from the Technical University of Munich in 2017. Currently, he is an assistant professor at the Department of Industrial Engineering (DII), University of Trento, Italy. Previously, he was an assistant professor at the University of Innsbruck and a post-doctoral researcher at the German Aerospace Center (DLR). He is an Associate Editor for the IEEE Robotics and Automation Letters and for The International Journal of Robotics Research. His research activities include robot learning, human-robot interaction, understanding and interpreting human activities. Webpage: https://matteosaveriano.weebly.com/
\end{IEEEbiography}


\vfill

\end{document}